\documentclass{article}

\usepackage{amsfonts}
\usepackage{amsmath}
\usepackage{booktabs} 
\usepackage{enumitem}
\usepackage{graphicx}
\usepackage{microtype}
\usepackage{placeins}
\usepackage{subfigure}
\usepackage{tabularx}
\usepackage{xspace}
\usepackage{stmaryrd}
\usepackage{textcomp}
\usepackage{wrapfig}
\usepackage[export]{adjustbox}

\usepackage[usenames,dvipsnames]{xcolor}
\definecolor{shadecolor}{gray}{0.9}

\usepackage{natbib}
\usepackage[colorlinks,linktoc=all,hypertexnames=false]{hyperref}
\usepackage[all]{hypcap}

\usepackage{xcolor}
\hypersetup{
  colorlinks,
  breaklinks=true,
  anchorcolor={blue!50!black},
  linkcolor={blue!50!black},
  citecolor={blue!50!black},
  urlcolor={blue!50!black}
}
\usepackage[nameinlink]{cleveref}
\creflabelformat{equation}{#2\textup{#1}#3}  

\newcommand{\ImNetA}{\textsc{ImageNet-A}\xspace}
\newcommand{\ImNetC}{\textsc{ImageNet-C}\xspace}
\newcommand{\ImNetR}{\textsc{ImageNet-R}\xspace}
\newcommand{\ImNetVtwo}{\textsc{ImageNetV2}\xspace}
\newcommand{\ImNetClean}{\textsc{ImageNet}\xspace}
\newcommand{\ImNettwentyonek}{\textsc{ImageNet-21k}\xspace}

\newcommand{\jft}{\textsc{JFT-300}\xspace}

\DeclareMathOperator*{\argmax}{arg\,max}

\usepackage[final]{neurips_2021}

\setcitestyle{authoryear,round,citesep={;},aysep={,},yysep={;}}

\usepackage[utf8]{inputenc} 
\usepackage[T1]{fontenc}    
\usepackage{url}            
\usepackage{booktabs}       
\usepackage{amsfonts}       
\usepackage{nicefrac}       
\usepackage{microtype}      
\usepackage{xcolor}         

\crefname{appendix}{Appendix}{Appendices}
\Crefname{appendix}{Appendix}{Appendices}

\title{Revisiting the Calibration of\\Modern Neural Networks}

\author{%
  Matthias Minderer \quad
  Josip Djolonga \quad
  Rob Romijnders \quad
  Frances Hubis \\
  \textbf{Xiaohua Zhai} \quad
  \textbf{Neil Houlsby} \quad
  \textbf{Dustin Tran} \quad
  \textbf{Mario Lucic}\vspace{1mm}\\
  Google Research, Brain Team\vspace{1mm}\\
  \texttt{\{mjlm, lucic\}@google.com} \\
}

\begin{document}

\maketitle

\begin{abstract}
Accurate estimation of predictive uncertainty (model calibration) is essential for the safe application of neural networks. 
Many instances of miscalibration in modern neural networks have been reported, suggesting a trend that newer, more accurate models produce poorly calibrated predictions. 
Here, we revisit this question for recent state-of-the-art image classification models.
We systematically relate model calibration and accuracy, and find that the most recent models, notably those not using convolutions, are among the best calibrated. 
Trends observed in prior model generations, such as decay of calibration with distribution shift or model size, are less pronounced in recent architectures.
We also show that model size and amount of pretraining do not fully explain these differences, suggesting that architecture is a major determinant of calibration properties.
\end{abstract}

\addtocontents{toc}{\protect\setcounter{tocdepth}{0}}

\section{Introduction}
Neural networks, especially vision models, are increasingly used in safety-critical applications such as autonomous driving~\citep{bojarski2016end}, medical diagnosis~\citep{esteva2017dermatologist,jiang2012calibrating}, and meteorological forecasting \citep{sonderby2020metnet}. For such applications, it is essential that model predictions are not just accurate, but also well calibrated. Model calibration refers to the accuracy with which the scores provided by the model reflect its predictive uncertainty. For example, in a medical application, we would like to defer images for which the model makes low-confidence predictions to a physician for review~\citep{kompa2021second}. Skipping human review due to confident, but incorrect, predictions, could have disastrous consequences.

While intense research and engineering effort has focused on improving the predictive accuracy of models, less attention has been given to model calibration.
In fact, over the last few years, there have been many reports that calibration of modern neural networks can be surprisingly poor, despite the advances in accuracy (e.g.\
\citealt{DBLP:conf/icml/GuoPSW17,DBLP:conf/nips/Lakshminarayanan17,DBLP:conf/nips/MalininG18,thulasidasan2019mixup,DBLP:conf/iclr/HendrycksMCZGL20,DBLP:conf/nips/SnoekOFLNSDRN19,DBLP:conf/nips/WenzelSTJ20,havasi2020training,DBLP:journals/corr/abs-2007-08792,leathart2020temporal}). Some works suggest a trend for larger, more accurate models to be worse calibrated \citep{DBLP:conf/icml/GuoPSW17}.

These concerns are more relevant than ever, since the architecture size, amount of training data, and computing power used by state-of-the-art models continue to increase. At the same time, rapid advances in model architecture \citep{tolstikhin2021mixer, dosovitskiy2020image} and training approaches \citep{DBLP:conf/icml/ChenK0H20, resnextwsl, radford2learning} raise the question whether past results on calibration, largely obtained on standard convolutional architectures, extend to current state-of-the-art models. Since model advances are quickly translated to real-world, safety-critical applications (e.g.~\citealt{mustafa2021supervised}), there is an urgent need to re-assess the calibration properties of current state-of-the-art models.

\textbf{Contributions.}\quad To address this need, we provide a systematic comparison of recent image classification models, relating their accuracy, calibration, and design features. We find that:
\begin{enumerate}[itemsep=0mm,topsep=0mm,parsep=0mm,leftmargin=2.5em]
    \item The best current models, including the non-convolutional MLP-Mixer \citep{tolstikhin2021mixer} and Vision Transformers \citep{dosovitskiy2020image}, are well calibrated compared to past models and their performance is more robust to distribution shift.
    \item In-distribution calibration slightly deteriorates with increasing model size, but this is outweighed by a simultaneous improvement in accuracy.
    \item Under distribution shift, calibration \emph{improves} with model size, reversing the trend seen in-distribution.
    \item Accuracy and calibration are correlated under distribution shift, such that optimizing for accuracy may also benefit calibration.
    \item Model size, pretraining duration, and pretraining dataset size cannot fully explain differences in calibration properties between model families.
\end{enumerate}
Our results suggest that further improvements in model accuracy will continue to benefit calibration. They also hint at architecture as an important determinant of model calibration. We provide code and a large dataset of calibration measurements, comprising 180 distinct models from 16 families, each evaluated on 79 ImageNet-scale datasets and 28 metric variants.\footnote{Available at \url{https://github.com/google-research/robustness_metrics/tree/master/robustness_metrics/projects/revisiting_calibration}.}
\section{Related Work}\label{sec:related_work}

\paragraph{Measures of model calibration.}
The losses that are commonly used to train classification models, such as cross-entropy and squared error, are proper scoring rules \citep{gneiting2007probabilistic} and are therefore guaranteed to yield perfectly calibrated models at their minimum---in the infinite-data limit. However, in practice, due to model mismatch and overfitting, even losses based on proper scoring rules may result in poor model calibration.
Miscalibration is commonly quantified in terms of Expected Calibration Error (ECE;~\citealt{naeini2015obtaining}), which measures the absolute difference between predictive confidence and accuracy. We focus on ECE because it is a widely used and accepted calibration metric. Nevertheless, it is well understood that estimating ECE accurately is difficult because estimators can be strongly biased and many estimator variants exist~\citep{DBLP:conf/cvpr/NixonDZJT19,roelofs2020mitigating,DBLP:conf/aistats/VaicenaviciusWA19,DBLP:journals/corr/abs-2006-12800}. \Cref{sec:pitfalls} discusses these issues and our approaches to mitigate them.

Alternatives to ECE include likelihood measures, Brier score~\citep{brier1950verification}, Bayesian methods~\citep{gelman2013bayesian}, and conformal prediction~\citep{shafer2008tutorial}. Further, model calibration  can be represented visually with reliability diagrams~\citep{degroot1983comparison}. \Cref{fig:alternative_calibration_metrics,app:alternative_calibration_metrics} provide likelihoods, Brier scores, and reliability diagrams for our main analyses.

\textbf{Empirical studies of model calibration.}\quad
There have been many recent empirical studies on the robustness (accuracy under distribution shift) of image classifiers \citep{geirhos2018imagenet,taori2020measuring,djolonga2020robustness,hendrycks2020many}.
Several works have also studied calibration. Most notable is \citet{DBLP:conf/icml/GuoPSW17}, who found that ``modern neural networks, unlike those from a decade ago, are poorly calibrated'', that larger networks tend to be calibrated worse, and that ``miscalibration worsen[s] even as classification error is reduced.'' Other works have corroborated some of these findings (e.g., \citealt{thulasidasan2019mixup,wen2020combining}). This line of work suggests a trend that larger models are worse calibrated, which would have major implications for research toward bigger models and datasets. We show that for more recent models, this trend is negligible in-distribution and in fact reverses under distribution shift.

\citet{DBLP:conf/nips/SnoekOFLNSDRN19} empirically study calibration under distribution shift and provide a large comparison of methods for improving calibration. They report that both accuracy and calibration deteriorate with distribution shift. While we observe the same trend, we find that the calibration of some recent model families decays so slowly under distribution shift that the decay in accuracy is likely more relevant in practice (\Cref{sec:imagenet_c}).

\citeauthor{DBLP:conf/nips/SnoekOFLNSDRN19} also find that, \emph{across methods for improving calibration}, improvements on in-distribution data do not necessarily translate to out-of-distribution data. This finding may suggest that there is little correlation between in-distribution and out-of-distribution calibration in general. However, our results show that, \emph{across model architectures}, the models with the best in-distribution calibration are also the best-calibrated on a range of out-of-distribution benchmarks. The important implication of this result is that designing models based on in-distribution performance likely also benefits their out-of-distribution performance.

\textbf{Improving calibration.}\quad
Many strategies have been proposed to improve model calibration
such as post-hoc rescaling of predictions \citep{DBLP:conf/icml/GuoPSW17},
averaging multiple predictions
\citep{DBLP:conf/nips/Lakshminarayanan17,wen2020batchensemble},
and
data augmentation \citep{thulasidasan2019mixup,wen2020combining}. Here, we focus on the intrinsic calibration properties of state-of-the-art model families, rather than methods to further improve calibration.

As a baseline on top of a model's intrinsic calibration properties, we study temperature scaling~\citep{DBLP:conf/icml/GuoPSW17}. It is effective in improving calibration and so simple that it can be applied in many cases at minimal additional cost, in contrast to many more sophisticated methods. Temperature scaling re-scales a model's logits by a single parameter, chosen to optimize the model's likelihood on a held-out portion of the training data. This temperature factor changes the model's \emph{confidence}, i.e., whether the model predictions are on average too certain (overconfident), optimally confident, or too uncertain (underconfident). The classification accuracy of the model is not affected by temperature scaling. A large fraction of model miscalibration is typically due to average over- or underconfidence, e.g.\ due to suboptimal training duration \citep{DBLP:conf/icml/GuoPSW17}. By normalizing a model's confidence, temperature scaling not only improves calibration, but also removes a primary confounder that can hide trends in calibration between models (see \Cref{sec:in_distribution,app:model_confidence}). Therefore, we study both unscaled and temperature-scaled predictions in the paper.
\section{Definitions and Notation}
\label{sec:measuring}

We consider the multi-class classification problem, as analyzed by \citet{brocker2009reliability}, where we observe a variable $X$ and predict a categorical variable $Y\in\{1, 2,\ldots, k\}$.
We model our predictor $f$ as a function that maps every input instance $X$ to a categorical distribution over $k$ labels,  represented using a vector $f(X)$ belonging to the $(k-1)$-dimensional simplex $\Delta=\{p \in[0,1]^k \mid \sum_{y=1}^k p_y=1\}$.

Intuitively, a model $f$ is well-calibrated if its output truthfully quantifies the predictive uncertainty.
For example, if we take all data points $x$ for which the model predicts $[f(x)]_y=0.3$, we expect $30\%$ of them to indeed take on the label $y$.
Formally, the model $f$ is said to be calibrated if \citep{brocker2009reliability}
\begin{equation}
  \forall p\in\Delta \colon P(Y = y \mid f(X) = p) = p_y.
\end{equation}
We will focus on a slightly weaker, but more practical condition, called top-label or argmax calibration \citep{DBLP:conf/nips/KumarLM19,DBLP:conf/icml/GuoPSW17}. This requires that the above holds only for the most likely label, i.e., $\forall p^*\in[0,1]$
\begin{equation}\label{eqn:argmax-calibration}
        P(Y \in \argmax p \mid \max f(X) = p^*) = p^*,
\end{equation}
where the $\max$ and $\argmax$ act coordinate-wise.

The most common measure of the degree of miscalibration is the \emph{Expected Calibration Error (ECE)}, which computes the expected disagreement between the two sides of eq.~\eqref{eqn:argmax-calibration}
\begin{equation}\label{eqn:ece}
  \mathbb{E}\big[| p^* - E[Y\in\argmax f(X) \mid \max f(X) = p^*|\big].
\end{equation}
Unfortunately, eq.~\eqref{eqn:ece} cannot be estimated without quantization as it conditions on a null event.
Hence, one typically first buckets the predictions into $m$ bins $B_1, \ldots, B_m$ based on their top predicted probability, and then takes the expectation over these buckets.
Namely, if we are given a set of $n$ i.i.d.\ samples $(x_1, y_1), \ldots, (x_n, y_n)$ distributed as $P(X,Y)$, then we assign each $j\in\{1,\ldots, n\}$ to a bucket $B_i$ based on $\max f(x_j)$.
Then, we compute in each bucket $B_i$ the  $\textrm{confidence}(B_i)=\frac{1}{|B_i|}\sum_{j\in B_i} \max f(x_j)$ and the $\textrm{accuracy}(B_i)=\frac{1}{|B_i|}\sum_{j\in B_i} \llbracket y_j\in \argmax f(x_j)\rrbracket$, where $\llbracket\cdot\rrbracket$ is the Iverson bracket.
Finally, we construct an estimator by taking the expectation over the bins
\begin{equation}
    \widehat{\textrm{ECE}} = \sum_{i=1}^{m} \frac{|B_i|}{n} \left|\text{accuracy}(B_i) - \text{confidence}(B_i)\right|.
\end{equation}
In \Cref{sec:pitfalls} we discuss the statistical properties of this estimator, possible pitfalls, and several mitigation strategies.

\section{Empirical Evaluation}

\begin{figure*}[t]
    \centering
    \includegraphics[width=\linewidth]{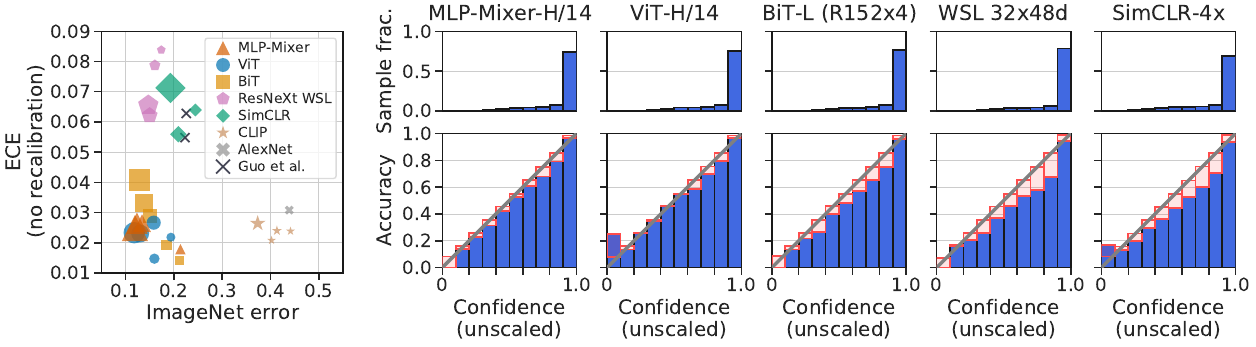}
    \caption{Some modern neural network families are both highly accurate and well-calibrated. Left: Expected calibration error (ECE) vs. classification error on \ImNetClean for state-of-the-art image classification models. Marker size indicates relative model size within its family. Points labeled ``Guo et al.'' are the values reported for DenseNet-161 and ResNet-152 in \citet{DBLP:conf/icml/GuoPSW17}. Right: Confidence distribution (top row) and reliability diagrams (bottom row) for some of the models.}
    \label{fig:figure_one}
\end{figure*}

\subsection{Experimental Setup}

\paragraph{Model families.}
In this study, we consider a range of recent and some historic state-of-the-art image classification models. Our selection of models covers convolutional and non-convolutional architectures, as well as supervised, weakly supervised, unsupervised and zero-shot training. We follow the original publications in naming the model variants within each family (e.g.\ different model sizes). See \Cref{app:model_overview} for a detailed description of all used models.
\begin{enumerate}[itemsep=0mm,topsep=0mm,parsep=0mm,leftmargin=2.5em]
    \item MLP-Mixer~\citep{tolstikhin2021mixer} is based exclusively on multi-layer perceptrons (MLPs) and is pre-trained on large supervised datasets.
    \item ViT~\citep{dosovitskiy2020image} processes images with a transformer architecture originally designed for language \citep{DBLP:conf/nips/VaswaniSPUJGKP17} and is also pre-trained on large supervised datasets.
    \item BiT~\citep{DBLP:conf/eccv/KolesnikovBZPYG20} is a ResNet-based architecture~\citep{DBLP:conf/eccv/HeZRS16}. It is also pre-trained on large supervised datasets.
    \item ResNext-WSL~\citep{resnextwsl} is based on the ResNeXt architecture and trained with weak supervision from billions of hashtags on social media images.
    \item SimCLR~\citep{DBLP:conf/icml/ChenK0H20} is a ResNet, pretrained with an unsupervised contrastive loss.
    \item CLIP~\citep{radford2learning} is pretrained on raw text and imagery using a contrastive loss.
    \item AlexNet~\citep{DBLP:conf/nips/KrizhevskySH12,krizhevsky2014one} was the first convolutional neural network to win the ImageNet challenge.

\end{enumerate}
All models are either trained or fine-tuned on the \ImNetClean training set, except for CLIP, which makes zero-shot predictions using \ImNetClean class names as queries.

\textbf{Datasets.}\quad
We evaluate accuracy and calibration on the \ImNetClean validation set and the following out-of-distribution benchmarks using the Robustness Metrics library \citep{djolonga2020robustness}:
\begin{enumerate}[itemsep=0mm,topsep=0mm,parsep=0mm,leftmargin=2.5em]
	\item \ImNetVtwo~\citep{DBLP:conf/icml/RechtRSS19} is a new \ImNetClean test set collected by closely following the original \ImNetClean labeling protocol.
	\item \ImNetC~\citep{DBLP:journals/corr/abs-1807-01697} consists of the images from \ImNetClean, modified with synthetic perturbations such as blur, pixelation, and compression artifacts at a range of severities.
	\item \ImNetR~\citep{hendrycks2020many} contains artificial renditions of \ImNetClean classes such as art, cartoons, drawings, sculptures, and others.
 	\item \ImNetA~\citep{DBLP:journals/corr/abs-1907-07174} contains images that are classified as belonging to \ImNetClean classes by humans, but adversarially selected to be hard to classify for a ResNet50 trained on \ImNetClean.
\end{enumerate}

For the post-hoc recalibration of models, we reserve 20\% of the \ImNetClean validation set (randomly sampled) for fitting the temperature scaling parameter. All reported metrics are computed on the remaining 80\% of the data. For evaluations on \ImNetC, we also exclude the 20\% of images that are based on the \ImNetClean images used for temperature scaling.

\textbf{Calibration metric.}\quad Throughout the paper, we estimate ECE using equal-mass binning and 100 bins. \Cref{app:ece_variants} shows that our results hold for other ECE variants and are consistent with the Brier score and model likelihood.

\subsection{In-Distribution Calibration}
\label{sec:in_distribution}

We begin by considering ECE on clean \ImNetClean images (referred to as in-distribution). \Cref{fig:figure_one} shows in-distribution ECE and reliability diagrams before any recalibration of the predicted probabilities. We find that several recent model families (MLP-Mixer, ViT, and BiT) are both highly accurate \emph{and} well-calibrated compared to prior models, such as AlexNet or the models studied by \citet{DBLP:conf/icml/GuoPSW17}. This suggests that there may be no continuing trend for highly accurate modern neural networks to be poorly calibrated, as suggested previously \citep{DBLP:conf/icml/GuoPSW17,DBLP:conf/nips/Lakshminarayanan17,DBLP:conf/nips/MalininG18,thulasidasan2019mixup,DBLP:conf/iclr/HendrycksMCZGL20,DBLP:conf/nips/SnoekOFLNSDRN19,DBLP:conf/nips/WenzelSTJ20,havasi2020training,DBLP:journals/corr/abs-2007-08792,leathart2020temporal}. In addition, we find that a recent zero-shot model, CLIP, is well-calibrated given its accuracy.

\begin{wrapfigure}{r}{0.5\textwidth}
    \capstart
    \centering
    \vspace{-4mm}
    \includegraphics[width=\linewidth]{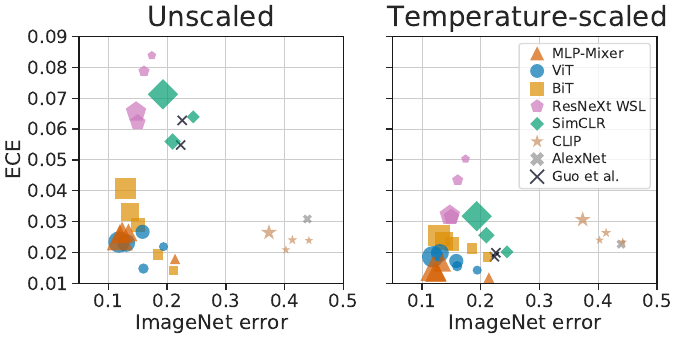}
    \caption{Temperature scaling reveals consistent properties of model families. Left: ECE vs. classification error as in \Cref{fig:figure_one}. Right: ECE after applying temperature scaling.}
    \label{fig:imagenet_clean_temp_scaled}
    \vspace{-4mm}
\end{wrapfigure}

\textbf{Temperature scaling reveals consistent properties of model families.}\quad
The poor calibration of past models can often be remedied by post-hoc recalibration such as temperature scaling \citep{DBLP:conf/icml/GuoPSW17}, which raises the question whether a difference between models remains after recalibration. We find that the most recent architectures are better calibrated than past models \emph{even after temperature scaling}  (\Cref{fig:imagenet_clean_temp_scaled}, right).

More generally, temperature scaling reveals consistent trends in the calibration properties between families that are obscured in the unscaled data by simple over- or underconfidence miscalibration. Before temperature scaling (\Cref{fig:imagenet_clean_temp_scaled}, left), several families overlap in their accuracy/calibration properties (MLP-Mixer, ViT, BiT). After temperature scaling (\Cref{fig:imagenet_clean_temp_scaled}, right), a clearer separation of families and consistent trends between accuracy and calibration within each family become apparent. Notably, temperature scaling reconciles our results for BiT (a ResNet architecture) with the results reported by \citeauthor{DBLP:conf/icml/GuoPSW17} for ResNets trained on \ImNetClean. Furthermore, models pretrained without additional labels (SimCLR) or with noisy labels (ResNeXt-WSL) tend to be calibrated worse for a given accuracy than ResNets trained with supervision (BiT and the models studied by \citeauthor{DBLP:conf/icml/GuoPSW17}). Finally, non-convolutional model families like MLP-Mixer and ViT can perform just as well, if not better, than convolutional ones.

\textbf{Differences between families are not explained by model size or pretraining amount.}\quad We next attempt to disentangle how the differences between model families affect their calibration properties. We focus on model size and amount of pretraining, both important trends in state of-the-art models.

We first consider model size. Prior work has suggested that larger neural networks are worse calibrated \citep{DBLP:conf/icml/GuoPSW17}. We also find that within most families, larger members tend to have higher calibration error (\Cref{fig:imagenet_clean_temp_scaled}, right). However, at the same time, larger models have consistently lower classification error. This means that each model family occupies a different Pareto set in the tradeoff between accuracy and calibration. For example, our results suggest that, at any given accuracy, ViT models are better calibrated than BiT models. Changing the size of a BiT model cannot move it into the Pareto set of ViT models. Model size can therefore not fully explain the intrinsic calibration differences between these model families.\footnote{This relationship between model size, accuracy and calibration holds for all families we study except ResNeXt-WSL, for which increasing model sizes improves \emph{both} accuracy \emph{and} calibration. While investigating this difference was out of the scope of this work, it may be a promising direction for future research.}

We next consider model pretraining. Many current state-of-the-art image models use transfer learning, in which a model is pre-trained on a large dataset and then fine-tuned to the task of interest \citep{DBLP:conf/eccv/KolesnikovBZPYG20,DBLP:conf/icml/ChenK0H20,DBLP:journals/corr/abs-1911-04252}. With transfer learning, large data sources can be exploited to train the model, even if little data are available for the final task. To test how the amount of pretraining affects calibration, we compare BiT models pretrained on \ImNetClean (1.3M images), \ImNettwentyonek (12.8M images), or \jft (300M images; \citealt{sun2017jft}).

\begin{wrapfigure}{r}{0.5\textwidth}
    \capstart
    \centering
    \vspace{-3mm}
    \includegraphics[width=\linewidth]{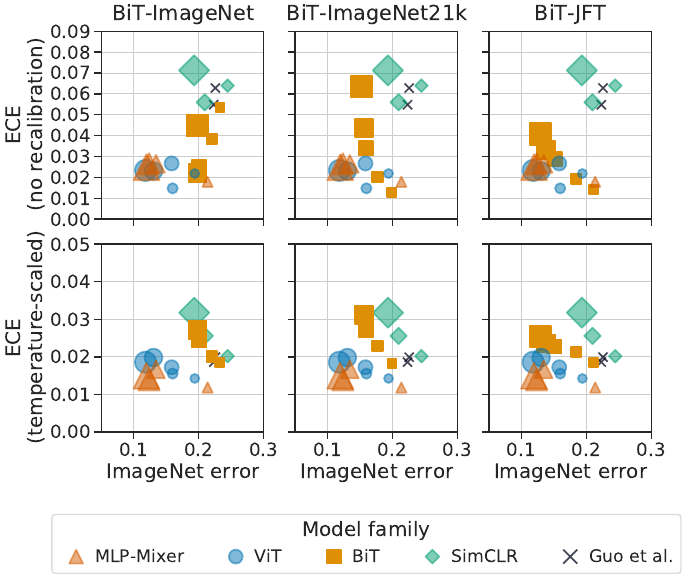}
    \caption{Family differences are not fully explained by the amount of pretraining. Each column shows ECE vs. classification error on ImageNet for BiT models pre-trained with a different dataset: \ImNetClean (1.3M images), \ImNettwentyonek (12.8M images), or \jft (300M images). The values for other models are provided for reference in light shading (same values as in \Cref{fig:imagenet_clean_temp_scaled}). Note how all BiT models remain in the same relative location between ViT and SimCLR across a 300-fold difference in pretraining data size.}
    \label{fig:imagenet_bit_pretrain}
    \vspace{-8mm}
\end{wrapfigure}

More pretraining data consistently increases accuracy, especially for larger models. It has no consistent effect on calibration (\Cref{fig:imagenet_bit_pretrain}). In particular, after temperature scaling, ECE is essentially unchanged across this 300-fold increase in pretraining dataset size (e.g.\ for BiT-R50x1 pretrained on \ImNetClean, \ImNettwentyonek and \jft, the ECEs are 0.0185, 0.0182, 0.0185, respectively; for BiT-R101x3, they are 0.0272, 0.0311, 0.0236; \Cref{fig:imagenet_bit_pretrain}, bottom). Therefore, regardless of the pretraining dataset, BiT always remains Pareto-dominant over SimCLR and Pareto-dominated by ViT and MLP-Mixer in our experiments.

The BiT models compared in \Cref{fig:imagenet_bit_pretrain} differ in both the amount of pretraining data and the duration of pretraining (see \citet{DBLP:conf/eccv/KolesnikovBZPYG20} for details). To further disentangle these variables, we trained BiT models on varying numbers of pretraining examples while holding the number of training steps constant, and vice versa. We find that pretraining dataset size has no significant effect on calibration, while pretraining duration only shifts the model within its accuracy/calibration Pareto set (longer-trained models are more accurate and worse calibrated; \Cref{fig:app_pretraining}). These results suggest that pretraining alone cannot explain the differences between model families that we observe.

In summary, our results show that some modern neural network families combine high accuracy and state-of-the-art calibration on in-distribution data, both before and after post-hoc recalibration by temperature scaling. In \Cref{fig:alternative_calibration_metrics,app:ece_variants,app:alternative_calibration_metrics}, we show that these results generally hold for other measures of model calibration (other ECE variants, Brier score, and model likelihood). Our experiments further suggest that model size and pretraining amount do not fully explain the intrinsic calibration differences between model families. Given that the best-calibrated families (MLP-Mixer and ViT) are non-convolutional, we speculate that model architecture, and in particular its spatial inductive bias, play an important role.

\subsection{Accuracy and Calibration Under Distribution Shift}\label{sec:imagenet_c}

\begin{figure*}[t]
    \centering
    \includegraphics[width=\linewidth]{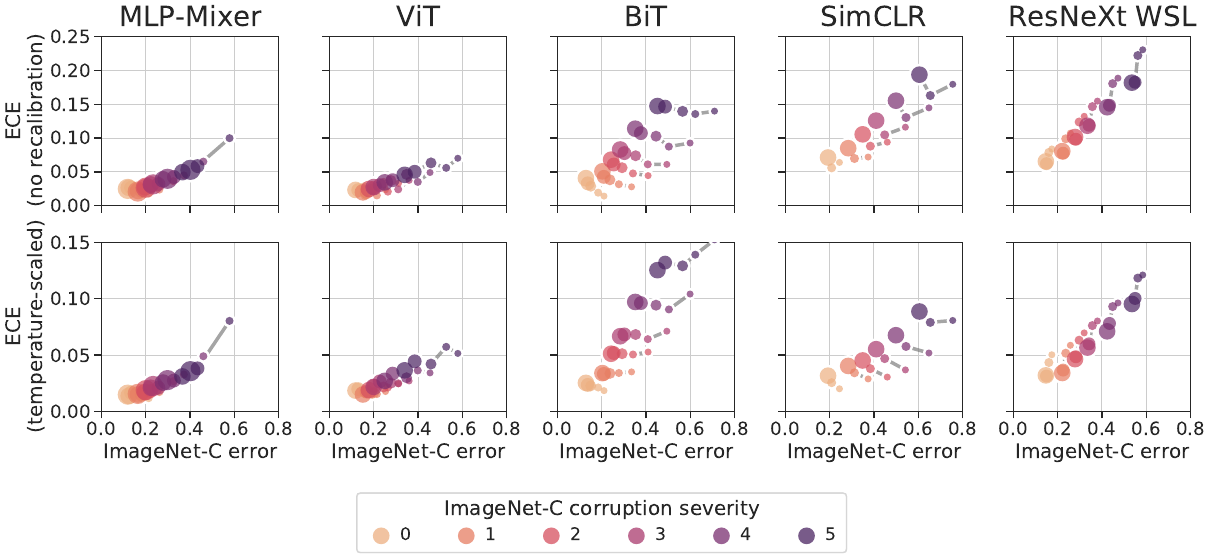}\vspace{-2mm}
    \caption{Calibration and accuracy on \ImNetC before (top) and after (bottom) temperature scaling on \ImNetClean. Severity~0 refers to the clean \ImNetClean test set; marker size indicates relative model size within its family (see \Cref{tab:model_training_overview} for model details). The calibration of some recent model families, e.g.\ MLP-Mixer and ViT, is more robust to distribution shift than past models.}\vspace{-2mm}
    \label{fig:imagenet_c}
\end{figure*}

For safety-critical applications, the model should produce reasonable uncertainty estimates not just in-distribution, but also under distribution shifts that were not anticipated at training time.
We first assess out-of-distribution calibration on the \ImNetC dataset, which consists of images that have been synthetically corrupted at five different severities.
As expected, both classification and calibration error generally increase with distribution shift (\Cref{fig:imagenet_c}; \citealt{DBLP:conf/nips/SnoekOFLNSDRN19,DBLP:journals/corr/abs-1807-01697}). Interestingly, this decay in calibration performance is slower for MLP-Mixer and ViT than for the other model families, both before and after temperature scaling.

\begin{wrapfigure}{r!}{0.5\textwidth}
    \capstart
    \centering
    \vspace{-3mm}
    \includegraphics[width=\linewidth]{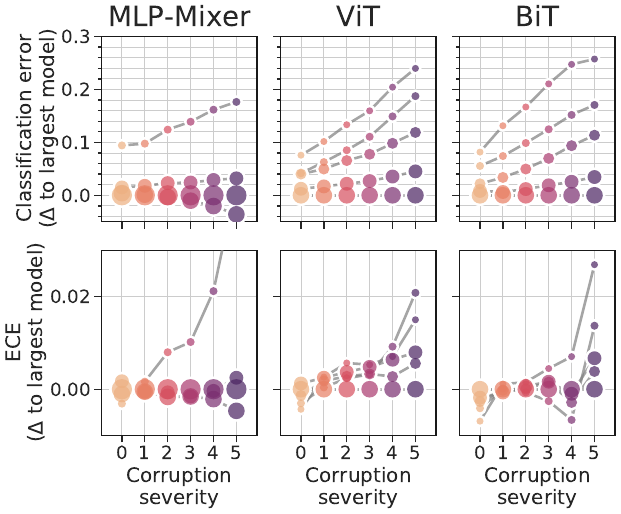}
    \caption{Classification error and ECE for the top three families on \ImNetC, relative to the largest model variant in each family. As distribution shift increases, both errors tend to increase more slowly for larger models. Also note that changes in ECE are much smaller than changes in classification error.}
    \label{fig:model_size_error_decay}
    \vspace{-5mm}
\end{wrapfigure}

Regarding the effect of model size on calibration, we observed some trend towards worse calibration of larger models on in-distribution data. However, the trend is reversed for most model families as we move out of distribution, especially after accounting for confidence bias by temperature scaling (note positive slope of the gray lines at high corruption severities in \Cref{fig:imagenet_c}, bottom row). In other words, the calibration of larger models is more robust to distribution shift (\Cref{fig:model_size_error_decay}).

\begin{figure*}[t]
    \capstart
    \centering
    \includegraphics[width=\linewidth]{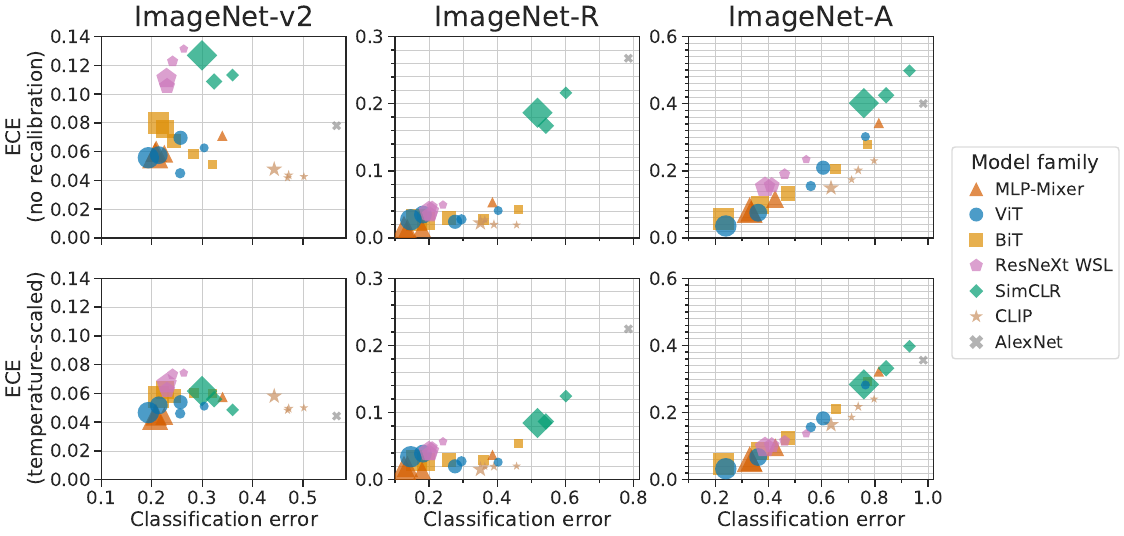}
    \caption{Calibration and accuracy before (top row) and after (bottom row) temperature scaling on out-of-distribution benchmarks. Marker size indicates relative model size within its family. \ImNetR and \ImNetA use a reduced subset of 200 classes; we follow the literature and select the subset of the model logits for these classes before evaluation. Out-of-distribution calibration tends to correlate with in-distribution calibration (\Cref{fig:figure_one}) and out-of-distribution accuracy.}
    \vspace{-4mm}
    \label{fig:ood_calibration}
\end{figure*}

We next consider to what degree the insights from in-distribution and \ImNetC calibration transfer to natural out-of-distribution data. Previous work on \ImNetC suggests that, when comparing recalibration methods, better in-distribution calibration and accuracy do not usually predict better calibration under distribution shift~\citep{DBLP:conf/nips/SnoekOFLNSDRN19}. Here, comparing model families, we find that the performance on several natural out-of-distribution datasets is largely consistent with that on \ImNetClean (\Cref{fig:ood_calibration}). In particular, models that are Pareto-optimal (i.e.\ no other model is both more accurate and better calibrated) on \ImNetClean remain Pareto-optimal on the OOD datasets. Further, we observe a strong correlation between accuracy and calibration on the OOD datasets. This relationship is consistent across models within a family and across datasets, over a wide range of accuracies (\Cref{fig:app_acc_calib_regression}).

These results suggest that larger and more accurate models, and in particular MLP-Mixer and ViT, can maintain their good in-distribution calibration even under severe distribution shifts. Based on the observed relationship between calibration and accuracy, we can reasonably hope that good calibration on in-distribution data (and anticipated distribution shifts) generally translates into good calibration on unanticipated out-of-distribution data, similar to what has been observed for accuracy \citep{djolonga2020robustness}.

\subsection{Relating Accuracy and Calibration Within Model Families}
\label{sec:selective_prediction}

Our data suggest that most model families lie on different Pareto sets in the accuracy/calibration space, which establishes a clear preference \emph{between} families. We next consider how to compare individual models \emph{within} a family (or more specifically, within a Pareto set), where one model is more accurate but worse calibrated, and the other is less accurate but better calibrated. Which model should a practitioner choose for a safety-critical application?

\begin{wrapfigure}{r!}{0.5\textwidth}
    \capstart
    \centering
    \vspace{-1mm}
    \includegraphics[width=\linewidth,trim={0mm 0mm 0mm 7mm},clip]{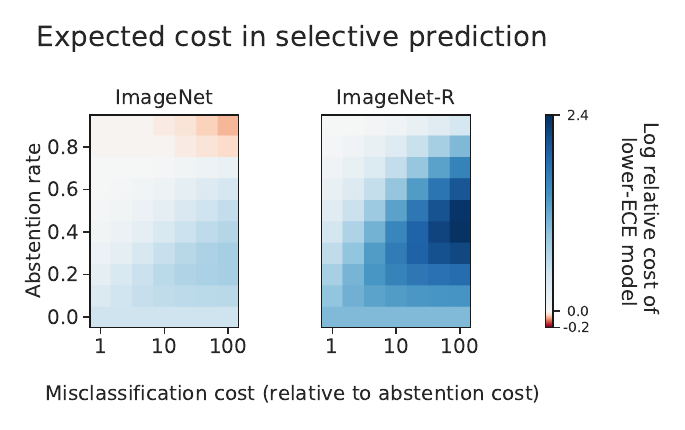}
    \vspace{-6mm}
    \caption{Relative cost of BiT-R152x4 and BiT-R50x1 models in a selective prediction scenario, computed as a combination of the misclassification and abstention costs at a given cost ratio (x-axis) and abstention rate (y-axis). Blue indicates regions where the higher-accuracy model (R152x4) achieves a lower cost than the better-calibrated model (R50x1). The accuracy advantage outweighs the calibration advantage for practical rejection rates, across all tested abstention costs.}
    \label{fig:abstention}
    \vspace{-4mm}
\end{wrapfigure}

The answer depends on the cost structure of the specific application \citep{DBLP:journals/jmlr/Hernandez-OralloFF12}. As an example, consider the scenario of \emph{selective prediction}, which is common in medical diagnosis. In this task, one can choose to ignore the model prediction (``abstain'') at a fixed cost if the prediction confidence is low, rather than risking a (more costly) prediction error.

\Cref{fig:abstention} compares the expected cost for two BiT variants, one with better with better classification error (R152x4, by 0.08), and one with better ECE (R50x1, by 0.009). For abstention rates up to 70\% (which covers most practical scenarios with abstention rates low enough for the model to be useful), the model with better accuracy has a lower overall cost than the model with better ECE. The same is true for all other model families we study (\Cref{app:abstention}). For these families and this cost scenario, a practitioner should therefore always choose the most accurate available model regardless of differences in calibration.
Ultimately, real-world cost structures are complex and may yield different results; \Cref{fig:abstention} presents one common scenario with downstream ramifications for the importance of the calibration differences compared to accuracy.

\section{Pitfalls and Limitations}\label{sec:pitfalls}
For this study, we approached calibration with a simple, practical question: \emph{Given two models, one more accurate and the other better calibrated, which should a practitioner choose?} While working towards answering this question, we encountered several pitfalls that complicate the interpretation of calibration results.

Measuring calibration is challenging, and while the quantity we want to estimate is well specified, the estimator itself can be biased.
There are two sources of bias: (i) from estimating ECE by binning, and (ii) from the finite sample size used to estimate the per-bin statistics.

\begin{figure*}[t]
    \centerline{\includegraphics[width=\textwidth]{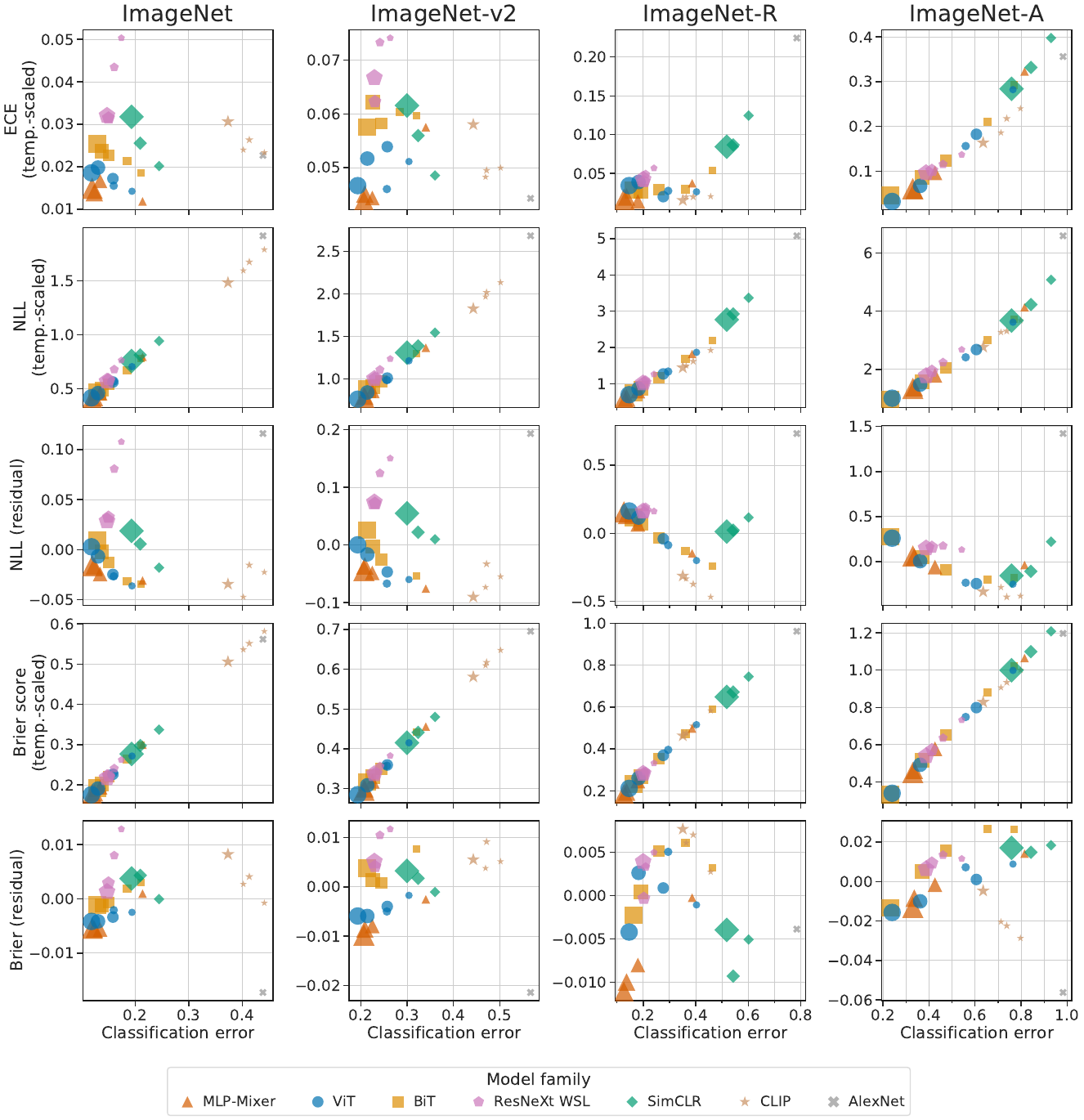}}
    \caption{Alternative calibration metrics: negative log-likelihood (NLL) and Brier score. For comparison, the first row shows ECE as in \Cref{fig:ood_calibration}. Since NLL, Brier score, and classification error are all highly correlated, we also provide the residuals of NLL and Brier score after regressing out classification error (third and fifth row). Specifically, we first fit a linear regression $y_i = \beta_0 + \beta_1 x_i$, where $x_i$ is the classification error and $y_i$ is the calibration measure of model $i$. We then report the residual $y_i - (\beta_0 + \beta_1 x_i)$ on the $y$-axis of the plots in the third and fifth row. The residuals show which models have better (or worse) NLL and Brier score than what can be expected from their accuracy alone. The relationships between model families are largely similar across all calibration metrics.}
    \label{fig:alternative_calibration_metrics}
\end{figure*}

\begin{wrapfigure}{r}{0.5\textwidth}
    \capstart
    \centering
    \includegraphics[width=\linewidth]{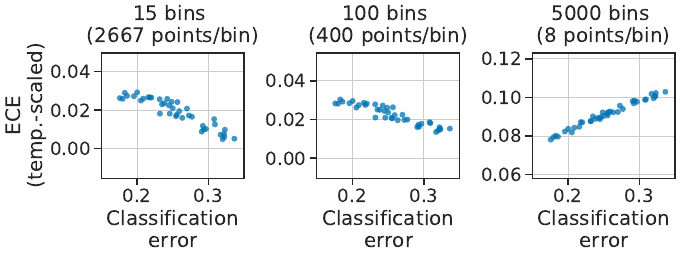}
    \caption{The effect of binning-induced bias in ECE depends on accuracy. Each dot represents a BiT ResNet model. Plotted models differ in model size, pretraining dataset size, and pretraining duration. All models are fine-tuned and evaluated on \ImNetClean. After temperature scaling, there is a near-linear relationship between ECE and classification error. However, whether this relationship is positive or negative depends on the number of bins used for estimating ECE. This effect is explained by an accuracy-dependent bias that increases with the number of bins.}
    \label{fig:bin_size_bias}
    \vspace{-4mm}
\end{wrapfigure}

The first of these biases is always negative \citep{DBLP:conf/nips/KumarLM19}, while the second one is always positive. Thus, the estimator can both under- and over-estimate the true value, and the magnitude of the bias can depend on multiple factors. In practice, this means that the ranking of models depends on which ECE variant is chosen to estimate calibration~\citep{DBLP:conf/cvpr/NixonDZJT19}. As we show below, this is especially problematic for the positive bias, because this bias \emph{depends on the accuracy of the model}. It is therefore possible to arrive at opposite conclusions about the relationship between accuracy and calibration, depending on the chosen bin size~(\Cref{fig:bin_size_bias}), especially when comparing models with widely varying accuracies.

Intuitively, a larger number of bins implies fewer points per bin and thus higher variance of the estimate of the model accuracy in each bin, which adds positive bias to the estimate of ECE. More formally, to estimate $\textrm{accuracy}(B_i)$ precisely, we need a number of samples inversely proportional to the standard deviation $\sqrt{p_i(1-p_i)}/|B_i|$, where $p_i$ is the expected accuracy in $B_i$.
This indicates that the bias would be smaller for models with extreme average accuracies (i.e.\ close to 0 or 1) and larger for models with an accuracy close to 0.5. A detailed analysis reveals additional effects that further reduce the bias for higher-accuracy models (\Cref{app:bias-proof}). In particular, if we estimate $(\mathbb{E}[\textrm{accuracy}(B_i)] - \mathbb{E}[\textrm{confidence}(B_i)])^2$ for any bin $i$ with $n_i$ samples (using the sample means of the confidences and the accuracies), the bias can be shown to be equal to (conditioning on $X\in B_i$ omitted for brevity)
\begin{equation}\label{eq:bias}
    \frac{1}{n_i}\big(\mathbb{V}[A] + \mathbb{V}[C] - 2 \textrm{Cov}[C, A]\big),
\end{equation}
where $A=\llbracket Y\in\argmax f(X)\rrbracket$ and $C=\max f(X)$.
Hence, from~\Cref{eq:bias} we can conclude that higher accuracy models have a lower bias not only due to higher accuracy (lower $\mathbb{V}[A]$), but also because their outputs correlate more with the correct label (higher covariance).

In addition to a careful choice of bin size, considering accuracy and calibration jointly mitigates this issue, because the Pareto-optimal models rarely change, even if the ranking based on ECE alone does (\Cref{app:ece_variants}). In \Cref{app:ece_variants,app:alternative_calibration_metrics}, we provide the main figures of the paper for other ECE variants (number of bins, binning scheme, normalization metric, top-label, all-label, class-wise).

Finally, metrics such as Brier score~\citep{brier1950verification} and likelihood provide alternative assessments of model calibration that do not require estimating expected calibration error. We find that the relationships between model families are consistent across ECE, NLL and Brier score (\Cref{fig:alternative_calibration_metrics}). In particular, the same models (specifically the largest MLP-Mixer and ViT variants) remain Pareto-optimal with respect to the calibration metric and classification error in most cases. The relationship between models is visualized especially clearly after regressing out from the calibration metrics their correlation with classification error (\Cref{fig:alternative_calibration_metrics}, third and fifth row).

\section{Conclusion}
We performed a large study of the calibration of recent state-of-the-art image models and its relationship with accuracy. We find that modern image models are well calibrated across distribution shifts despite being designed with a focus on accuracy. Our results suggests that there is no general trend for recent or highly accurate neural networks to be poorly calibrated compared to older or less accurate models.

Our experiments suggest that simple dimensions such as model size and pretraining amount do not fully account for the performance differences between families, pointing towards architecture as a major determinant of calibration. Of particular note is the finding that MLP-Mixer and Vision Transformers---two recent architectures that are not based on convolutions---are among the best-calibrated models both in-distribution and out-of-distribution. Self-attention (which Vision Transformers employ heavily) has been shown previously to be beneficial for certain kinds of out-of-distribution robustness \citep{hendrycks2020many}. Our work now hints at calibration benefits of non-convolutional architectures more broadly, for certain kinds of distribution shift. Further work on the influence of architectural inductive biases on calibration and out-of-distribution robustness will be necessary to tell whether these results generalize. If so, they may further hasten the end of the convolutional era in computer vision.

\begin{ack}
We thank Carlos Riquelme and Balaji Lakshminarayanan for valuable comments on the manuscript. The authors declare no competing interests.
\end{ack}

\bibliography{bibliography}
\bibliographystyle{icml2021}
\clearpage

\clearpage
\appendix

\addtocontents{toc}{\protect\setcounter{tocdepth}{2}}

\newcommand{\secsummary}[1]{\addtocontents{toc}{\vspace{-0.3em}\setlength{\leftskip}{5.4mm}\textit{#1}\par}}
\newcommand{\subsecsummary}[1]{\addtocontents{toc}{\vspace{-0.3em}\setlength{\leftskip}{13.4mm}\textit{#1}\par}}

{\Large{\textbf{Appendix}}}\vspace{5mm}

The Appendix is structured as follows:

\makeatletter
\renewcommand\tableofcontents{%
    \@starttoc{toc}%
}
\makeatother

\makeatletter
\renewcommand\@dotsep{200}
\makeatother

\tableofcontents

\clearpage
\section{Models and Datasets}
\secsummary{Details and references for the models and datasets used in this work.}

\subsection{Models}\label{app:model_overview}

Table \ref{tab:model_training_overview} provides an overview of the models used in this study. Model names link to the used checkpoints, where available.

\begin{table*}[ht]
    \addtolength{\tabcolsep}{-3pt}  
    \centering
     \begin{tabular}{l l l r r r} 
     \toprule 
Model name                             & Reference                         & Variant                                            & Parameters\\
\midrule
\href{https://pytorch.org/hub/pytorch_vision_alexnet/}{AlexNet}                 & \citet{DBLP:conf/nips/KrizhevskySH12}       & 8 layers                    & 62.4M \\
BiT-L (R50-x1)                                                                  & \citet{DBLP:conf/eccv/KolesnikovBZPYG20}    & ResNet50, 1$\times$ width   & 25.5M \\
BiT-L (R101-x1)                                                                 & \citet{DBLP:conf/eccv/KolesnikovBZPYG20}    & ResNet101, 1$\times$ width  & 44.5M \\
BiT-L (R50-x3)                                                                  & \citet{DBLP:conf/eccv/KolesnikovBZPYG20}    & ResNet50, 3$\times$ width   & 217.3M \\
BiT-L (R101-x3)                                                                 & \citet{DBLP:conf/eccv/KolesnikovBZPYG20}    & ResNet101, 3$\times$ width  & 387.9M \\
BiT-L (R152-x4)                                                                 & \citet{DBLP:conf/eccv/KolesnikovBZPYG20}    & ResNet154, 4$\times$ width  & 936.5M \\
\href{https://github.com/openai/CLIP}{CLIP}                                     & \citet{radford2learning}                    & ResNet50-based              & 25.5M \\
\href{https://github.com/openai/CLIP}{CLIP}                                     & \citet{radford2learning}                    & ViT-B32-based               & 88.3M \\
EfficientNet-NS (B1)                                                            & \citet{DBLP:journals/corr/abs-1911-04252}   & 18 layers, 1$\times$ width  & 7.9M \\
EfficientNet-NS (B3)                                                            & \citet{DBLP:journals/corr/abs-1911-04252}   & 31 layers, 1$\times$ width  & 12.3M \\
EfficientNet-NS (B5)                                                            & \citet{DBLP:journals/corr/abs-1911-04252}   & 45 layers, 2$\times$ width  & 30.6M \\
EfficientNet-NS (B7)                                                            & \citet{DBLP:journals/corr/abs-1911-04252}   & 64 layers, 2$\times$ width  & 66.7M \\
Mixer (B)                                                                       & \citet{tolstikhin2021mixer}                 & B/16, JFT-300m              & 59.9M \\
Mixer (L)                                                                       & \citet{tolstikhin2021mixer}                 & L/16, JFT-300m              & 280.5M \\
Mixer (H)                                                                       & \citet{tolstikhin2021mixer}                 & H/14, JFT-300m              & 589.7M \\
Mixer (H)                                                                       & \citet{tolstikhin2021mixer}                 & H/14, JFT-2.5b              & 589.7M \\
\href{https://pytorch.org/hub/facebookresearch_WSL-Images_resnext/}{ResNeXt-WSL}& \citet{resnextwsl}                          & ResNeXt 101, 32x8d          & 88M \\
\href{https://pytorch.org/hub/facebookresearch_WSL-Images_resnext/}{ResNeXt-WSL}& \citet{resnextwsl}                          & ResNeXt 101, 32x16d         & 193M \\
\href{https://pytorch.org/hub/facebookresearch_WSL-Images_resnext/}{ResNeXt-WSL}& \citet{resnextwsl}                          & ResNeXt 101, 32x32d         & 466M \\
\href{https://pytorch.org/hub/facebookresearch_WSL-Images_resnext/}{ResNeXt-WSL}& \citet{resnextwsl}                          & ResNeXt 101, 32x48d         & 829M \\
SimCLR (1x)                                                                     & \citet{DBLP:conf/icml/ChenK0H20}            & ResNet50, 1$\times$ width   & 25.6M \\
SimCLR (2x)                                                                     & \citet{DBLP:conf/icml/ChenK0H20}            & ResNet50, 2$\times$ width   & 98.1M \\
SimCLR (4x)                                                                     & \citet{DBLP:conf/icml/ChenK0H20}            & ResNet50, 4$\times$ width   & 383.8M \\
\href{https://github.com/google-research/vision_transformer}{ViT (B)}           & \citet{dosovitskiy2020image}                & B/32                        & 88.3M \\
\href{https://github.com/google-research/vision_transformer}{ViT (B)}           & \citet{dosovitskiy2020image}                & B/16                        & 86.9M \\
\href{https://github.com/google-research/vision_transformer}{ViT (L)}           & \citet{dosovitskiy2020image}                & L/32                        & 306.6M \\
\href{https://github.com/google-research/vision_transformer}{ViT (L)}           & \citet{dosovitskiy2020image}                & L/16                        & 304.7M \\
\href{https://github.com/google-research/vision_transformer}{ViT (H)}           & \citet{dosovitskiy2020image}                & H/14                        & 633.2M \\
     \bottomrule
     \end{tabular}%
    \caption{\label{tab:model_training_overview} Overview of models used in this study. Per model family, the rows are sorted by increasing marker size in \Cref{fig:figure_one} (i.e. approximate relative model size in terms of pretraining compute). We chose a qualitative scale to indicate model size because quantitative measures such as the number of parameters do not always reflect the representational power of a model. For example, ViT-B/16 has slightly fewer parameters than ViT-B/32 but requires more compute and is a more powerful model.}
\end{table*}

\subsection{Datasets} \label{app:datasets}

We evaluate accuracy and calibration the following benchmark datasets:
\begin{enumerate}[itemsep=0.25em,topsep=0em,parsep=0em,leftmargin=2.5em]
    \item \ImNetClean~\citep{imagenet2009} refers to the ILSVRC-2012 variant of the ImageNet database, a dataset of images of 1\,000 diverse object classes. For evaluation, we use 40\,000 images randomly sampled from the public validation set. We reserve the remaining 10\,000 images for fitting the temperature scaling parameter.

	\item \ImNetVtwo~\citep{DBLP:conf/icml/RechtRSS19} is a new \ImNetClean test set collected by closely following the original \ImNetClean labeling protocol. The dataset contains 10\,000 images.
	
	\item \ImNetC~\citep{DBLP:journals/corr/abs-1807-01697} consists of the images from \ImNetClean, modified with synthetic perturbations such as blur, pixelation, and compression artifacts at a range of severities. The dataset includes 15 perturbations at 5 severities each, for a total of 75 datasets. For evaluation, we use the 40\,000 images that were not derived from the \ImNetClean images we used for temperature scaling.
	
	\item \ImNetR~\citep{hendrycks2020many} contains artificial renditions of \ImNetClean classes such as art, cartoons, drawings, sculptures, and others. The dataset has 30\,000 images of 200 classes. Following \citeauthor{hendrycks2020many}, we sub-select the model logits for the 200 classes before computing accuracy and calibration metrics.
	
 	\item \ImNetA~\citep{DBLP:journals/corr/abs-1907-07174} contains images that are classified as belonging to \ImNetClean classes by humans, but adversarially selected to be hard to classify by a ResNet50 trained on \ImNetClean. The dataset has 7\,500 samples of 200 classes. As for \ImNetR, we sub-select the logits for the 200 classes before computing accuracy and calibration metrics.
\end{enumerate}

In addition, the following datasets are used for pretraining as described in the text:
\begin{enumerate}[itemsep=0.25em,topsep=0em,parsep=0em,leftmargin=2.5em]
	\item \ImNettwentyonek~\citep{imagenet2009} refers to the full variant of the ImageNet database. It contains 14.2 million images of 21\,000 object classes, organized by the WordNet hierarchy. Each image may have several labels.
	\item \jft~\citep{sun2017jft} consists of approximately 300 million images, with 1.26 labels per image on average. The labels are organized into a hierarchy of 18\,291 classes. 
\end{enumerate}

\section{Supplementary Analyses}
\subsection{Fine-grained Analysis of Pretraining}
\subsecsummary{Analyzes how the number of pretraining steps and the pretraining dataset size affect calibration and accuracy.}

\begin{figure}[h]    
    \centering
    \vspace{5mm}
    \includegraphics[width=0.75\textwidth]{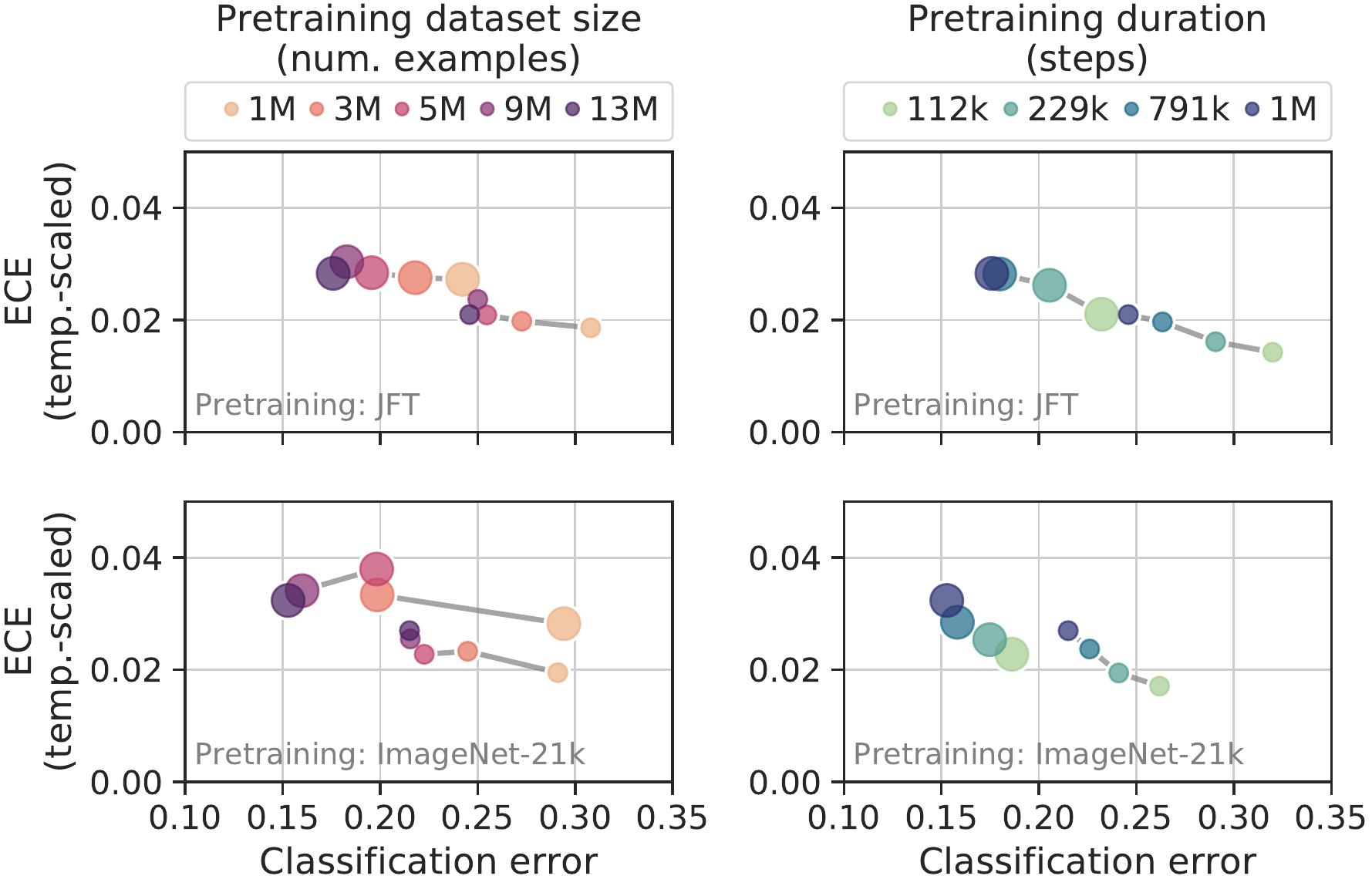}
    \caption{Effect of pretraining dataset size and duration on calibration. Larger dots indicate BiT-R101x3, smaller dots indicate BiT-R50x1. The pretraining datasets are subsampled from \jft (top) or \ImNettwentyonek (a larger variant of \ImNetClean; bottom). Classification error is on \ImNetClean after fine-tuning.}
    \label{fig:app_pretraining}
\end{figure}

\Cref{sec:in_distribution} and \Cref{fig:imagenet_bit_pretrain} discuss the effect of the amount of pretraining on accuracy and calibration by comparing models pretrained on three different datasets. \Cref{fig:app_pretraining} provides a more fine-grained analysis. We pretrained BiT models with varying dataset sizes or number of pretraining steps, while holding the other constant. Learning rate schedules were appropriately adapted to the number of steps, i.e. a separate model was trained with a full schedule for each condition, rather than comparing different checkpoints from the same training run. After pretraining, all models were finetuned on \ImNetClean as in \citet{DBLP:conf/eccv/KolesnikovBZPYG20}. 

We find that pretraining dataset size has little consistent effect on calibration error (\Cref{fig:app_pretraining}, left). Longer pretraining causes a slight increase in calibration error, but also decreases classification error (\Cref{fig:app_pretraining}, right).

\subsection{Correlation Between Calibration and Accuracy} \label{app:acc_cal_regression}
\subsecsummary{Shows the relationship between calibration and accuracy, separated by model family and without averaging ImageNet-C datasets and severities.}

\begin{figure}[h]
    \centering    
    \includegraphics[width=\textwidth]{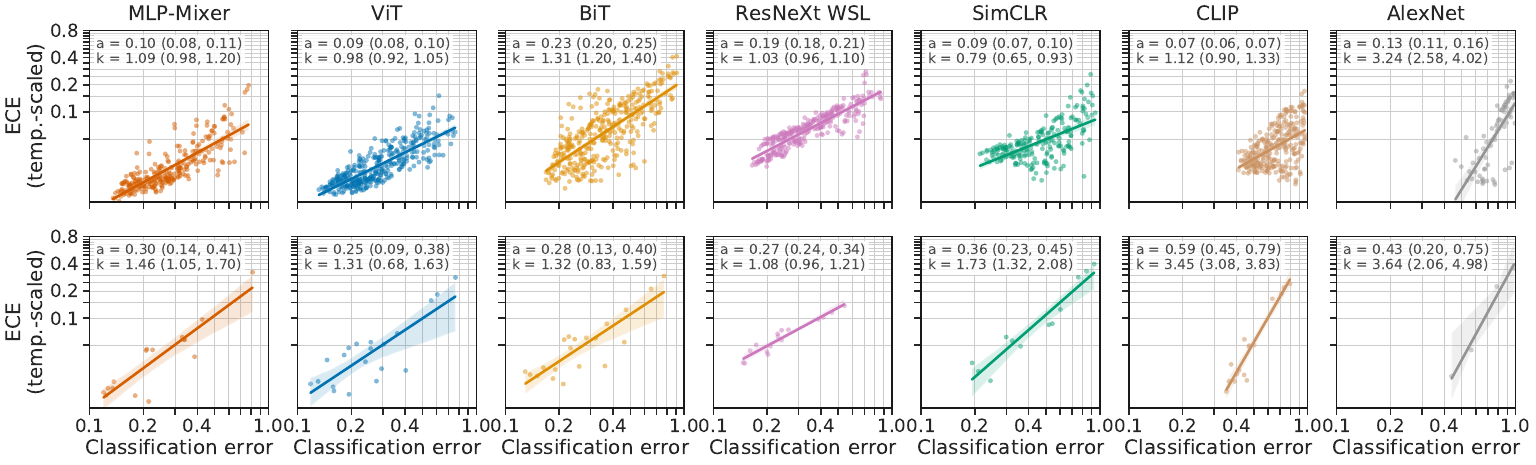}
    \caption{Correlation between ECE and classification error. Each dot represents a different combination of model variant and evaluation dataset (top: \ImNetC variants \emph{without} averaging corruptions and severities; bottom: allo other datasets). Lines show power laws of the form $y=ax^k$ where $x$ is classification error and $y$ is ECE. Range in parentheses indicates the 95\% confidence interval by bootstrap.}
    \label{fig:app_acc_calib_regression}
\end{figure}

\Cref{fig:imagenet_c,fig:ood_calibration} show that, across a sufficiently large range of distribution shift, calibration error and classification error are correlated. \Cref{fig:app_acc_calib_regression} illustrates this correlation for each model family across model variants and datasets. 

In general, it is expected that calibration error and classification error are correlated to some degree due to noise in the model predictions, since adding random noise to the model confidence score would increase both calibration and classification error. Indeed, all model families show a strong positive correlation between calibration and classification error. However, there are consistent differences between model families, reflecting their intrinsic calibration properties. The relationship can be remarkably strong and lawful. For example, a simple power law of the form $y=ax^k$ (where $x$ is classification error and $y$ is ECE) provides a good fit for some model families (e.g. ResNeXt WSL; \Cref{fig:app_acc_calib_regression}). The parameters of the fit provide a quantitative description of the intrinsic calibration properties of a model family that goes beyond ECE on a specific dataset.

\subsection{Contribution of Accuracy and Calibration to Decision Cost} \label{app:abstention}
\subsecsummary{Compares selective prediction cost of additional model pairs.}

\begin{figure}[!h]
    \centering
    \includegraphics[width=\linewidth,trim={0mm 0mm 0mm 7mm},clip]{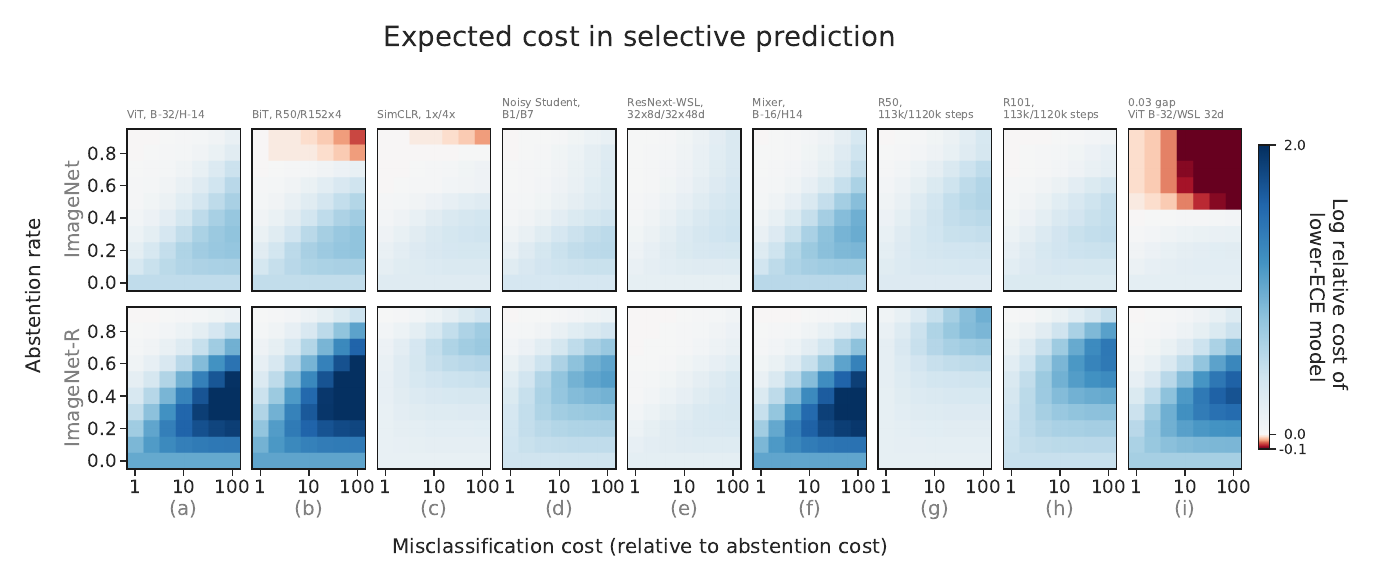}\vspace{-4mm}
    \caption{Relative impact of accuracy and calibration in a selective prediction scenario. Each heatmap compares two models and shows the relative cost of the better-calibrated (lower-ECE) model with respect to the other model. Total cost is computed as a linear combination of the misclassification cost and abstention cost at a given cost ratio (x-axis) and abstention rate (y-axis). Compared model pairs are indicated above each column. The top row shows \ImNetClean, the bottom row \ImNetR. In most scenarios, the higher-accuracy model is preferred over the better-calibrated model (blue regions). Only in a few cases and at very high abstention rates does the difference in calibration outweigh the difference in accuracy (red regions). In other words, for practical abstention rates and across a wide range of abstention costs, the accuracy advantage outweighs the calibration advantage.}
    \label{fig:abstention_appendix}
\end{figure}

In \Cref{sec:selective_prediction}, we use a selective prediction task as a practical scenario in which we can quantify the relative impact of accuracy and calibration on the ultimate decision cost incurred by a model user. In this task, the user can either accept a model prediction and incur a misclassification cost if the prediction is wrong, or reject (abstain from) the prediction and incur an abstention cost (which is independent of whether the model prediction would have been correct). This decision is made based on the model's confidence. The total cost therefore depends on both the accuracy and the calibration of the model. A concrete example is a medical diagnosis task in which we can choose to use the model's diagnosis as-is, or refer the case to a human for review. \Cref{fig:abstention_appendix} shows cost planes for eight model pairs. 

First, we compare models \textbf{from the same family} (\Cref{fig:abstention_appendix}, a--e). In the blue regions, the relative cost for the model with higher accuracy (always the larger model) is lower (better); in red regions, the relative cost of the model with lower accuracy (always the smaller model) is lower (better). For most models and for most practical cost settings, the higher accuracy model is preferred over the better calibrated model. In the ViT family, for example, the bigger model has 0.076 lower classification error (0.194 vs. 0.118) and 0.007 higher ECE (0.017 vs. 0.01). For these models, the cost analysis shows that the difference in classification error outweighs the difference in ECE across all tested misclassification costs and abstention rates.

Next, we compare models \textbf{pretrained for a different number of steps} (same models as used in \Cref{fig:app_pretraining}) and provide the results in \Cref{fig:abstention_appendix}, f--g. Again, the models with lower classification error (e.g. for R101x3, 0.176 vs 0.232 in favor of the longer-trained model) reach a lower total cost than the models with lower ECE (e.g. for R101x3, 0.019 vs 0.028 in favor of the shorter-trained model).

Finally, we compare models which attain \textbf{similar classification error and ECE difference} (\Cref{fig:abstention_appendix}, h). In particular, we compare ViT-B/32 and ResNeXt-WSL 32d models. The latter model has 0.03 lower (better) classification error while the ViT model has 0.03 lower (better) ECE. Again, for most practical cost settings, the model with better accuracy has lower cost than (is preferred over) the model with better ECE.

\section{Sampling Bias for $\ell_2$-ECE}\label{app:bias-proof}
\secsummary{Provides further analysis of the bias of ECE and a proof that the bias depends on accuracy.}
\newtheorem{lemma}{Lemma}

In \Cref{sec:pitfalls}, we hinted at the fact that the bias of the ECE estimator depends on the model accuracy.
Here, we expand on \Cref{eq:bias} and fully derive the bias for a variant of the ECE score, when we take the squared instead of the absolute differences in each bucket for tractability.

\begin{lemma}
Define the random variables $A=Y\in\argmax f(X)$ and $C=\max f(X)$, consider the squared ECE metric
\[  \textrm{ECE}_2 = \sum_{i=1}^m P(X\in B_i)(\textrm{accuracy}(B_i) - \textrm{confidence}(B_i))^2, \]
where the $B_i$ represent the $m$ disjoint buckets.
If we estimate the the per-bin statistics using their sample means, the statistical bias is equal to
\[
    \sum_i \frac{1}{n}\mathbb{V}[C-A \mid X\in B_i] = \frac{1}{n}\sum_i (\alpha_i (1-\alpha_i)(1 - \delta_i) + \mathbb{V}[C \mid X\in B_i]),
\]
where $\alpha_i$ is the accuracy in bucket $B_i$ and $\delta_i$ is the expected difference in the confidences of the correct and incorrect predictions. 
\end{lemma}

We assume that the buckets are fixed, s.t., there are $n_i$ points in bucket $B_i$, and a total of $n=\sum_i n_i$ points (we will take the expectation over $n_i$).
We introduce two random variables --- the model confidence by $C=\max f(X)$ and the corresponding true/false indicator by $A=\llbracket Y\in\argmax f(X) \rrbracket$.
For each realization $(x_j, y_j)$ we denote by $c_j$ and $a_j$ the corresponding values.
We further define for each bucket $B_i$
\begin{itemize}
    \item $\alpha_i=\mathbb{E}[A \mid X\in B_i]$, the accuracy in bucket $B_i$.
    \item $\gamma_i=\mathbb{E}[C \mid X\in B_i]$, the expected confidence in bucket $B_i$.
    \item  $\delta_i=\mathbb{E}[C \mid A=1, X\in B_i] - \mathbb{E}[C \mid A=0, X\in B_i]$, the confidence difference for the correct and wrong predictions.
    \item $\bar{c}_i=\sum_{c\in B_i} c / n_i$ , the sample average confidence in bucket $B_i$.
    \item $\bar{a}_i=\sum_{a\in B_i} a / n_i$ , the sample average accuracy in bucket $B_i$.
\end{itemize}

We consider the squared ECE $\ell_2$ loss, which after bucketing is equal to
\begin{align*}
    S^2 &= \sum_i P(X\in B_i) (
        \underbrace{\mathbb{E}[C \mid X\in B_i]}_{\gamma_i} -
        \underbrace{\mathbb{E}[A \mid X\in B_i]}_{\alpha_i})^2, \textrm{ and the corresponding sample estimate is} \\
    \hat{S}^2 &= \sum_i \frac{n_i}{n} (\frac{1}{n_i}\sum_{i\in B_i} c_i - \frac{1}{n_i}\sum_{j\in B_i} a_j)^2.
\end{align*}
The goal is to understand the bias $\hat{S}^2 - S^2$. Note that
\[
    \mathbb{E}[(\bar{c}_i - \bar{a}_i)^2 \mid n_i] = (\gamma_i - \alpha_i)^2 + \mathbb{V}[\bar{c}_i - \bar{a}_i \mid n_i] = (\gamma_i - \alpha_i)^2 + \frac{1}{n_i}\mathbb{V}[C - A \mid X \in B_i].
\]
We further have
\[
\mathbb{V}[C - A \mid X\in B_i] = \mathbb{V}[C  \mid X\in B_i] + \mathbb{V}[A \mid X\in B_i] - 2 \textrm{Cov}[C, A \mid X\in B_i].
\]
Hence, we have that
\begin{align*}
\mathbb{E}[\hat{S}^2] &= \mathbb{E}[\sum_i \frac{n_i}{n} (\frac{1}{n_i}\sum_{i\in B_i} c_i - \frac{1}{n_i}\sum_{i\in B_i} a_i)^2] \\
 &= \mathbb{E}[\mathbb{E}[\sum_i \frac{n_i}{n} (\frac{1}{n_i}\sum_{i\in B_i} c_i - \frac{1}{n_i}\sum_{i\in B_i} a_i)^2 \mid n_i]] \\
&= \mathbb{E}[\sum_i \frac{n_i}{n} \big( (\gamma_i - \alpha_i )^2 + \frac{1}{n_i}\mathbb{V}[C - A \mid X\in B_i] \big) ] \\
&= \sum_i(\gamma_i - \alpha_i )^2  \underbrace{\mathbb{E}[\frac{n_i}{n}]}_{P(X\in B_i)} + \sum_i\frac{1}{n}\mathbb{V}[C - A \mid X\in B_i] \\
&= S^2 + \underbrace{\frac{1}{n}\mathbb{V}[C - A \mid X\in B_i]}_{\textrm{bias}}.
\end{align*}

We can decompose the covariance as follows (see \href{https://stats.stackexchange.com/questions/50817/covariance-of-binary-and-continuous-variable/69698#69698}{this MathOverflow} answer), using the fact that $A$ is binary:
\begin{align*}
\textrm{Cov}[C, A \mid X \in G_i]
&= \mathbb{V}[\alpha_i]\delta_i,
\end{align*}

Here $\delta_i$ is defined as $\mathbb{E}[C \mid A=1, X \in B_i] - \mathbb{E}[C \mid A=0, X \in B_i]$. Now the total bias can be written as
\[
    \textrm{bias} = \frac{1}{n}\sum_i \alpha_i (1-\alpha_i)(1 - 2\delta_i) + \mathbb{V}[C \mid X\in B_i].
\]
Note that $\partial_{\alpha_i}\textrm{bias} = (1 - 2 \alpha_i)(1 - 2 \delta_i)$, which is negative when $\alpha_j > 1/2$, if we assume we have enough bins so that $\delta_j < 1/2$. Hence, for models that have at least $50\%$ top-1 accuracy, increasing the accuracy reduces the bias.

\clearpage
\section{Model Confidence} \label{app:model_confidence}
\secsummary{Discusses factors influencing model over- and underconfidence, and how temperature scaling removes confidence miscalibration as a confounding variable when comparing models.}

\begin{figure}[b!]
    \centering
    \includegraphics[width=\linewidth]{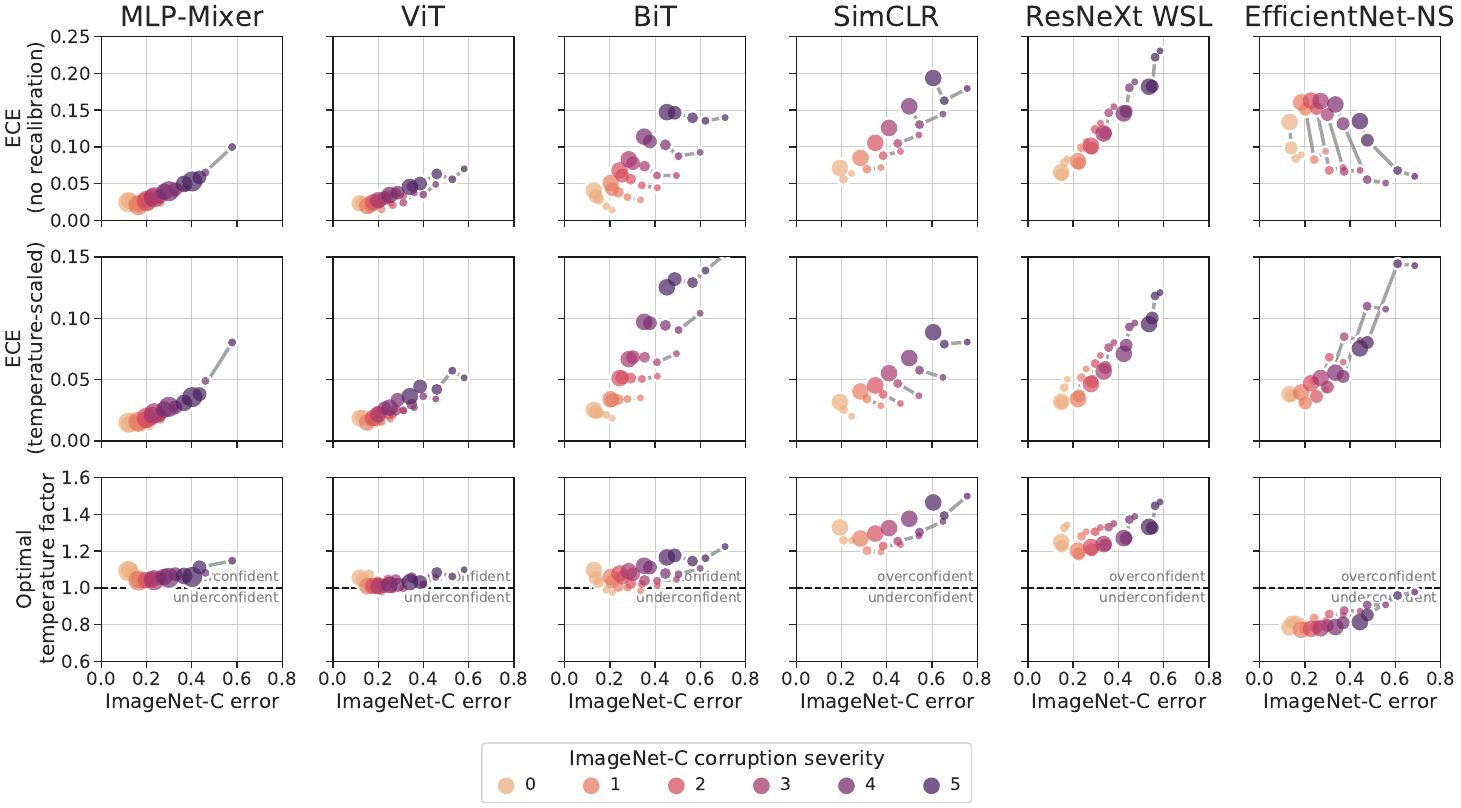}
    \caption{Related to \Cref{fig:imagenet_c}. Calibration and accuracy on \ImNetC. Here, the model confidence is shown in the third row (top two rows are identical to \Cref{fig:imagenet_c}). Model confidence is quantified in terms of the temperature scaling factor by which the logits of the unscaled model would have to be multiplied to provide optimal confidence for a given dataset. Values above 1 mean that the unscaled model is overconfident on the given dataset, and below 1, that the unscaled model is underconfident.}
    \label{fig:app_imagenet_c_with_confidence}
\end{figure}

An important aspect of the calibration of a model is its average \emph{confidence}, i.e. the systematic bias of the model's scores to be too high (overconfident) or too low (underconfident) compared to the true accuracy. A large fraction of the miscalibration of modern neural networks is typically due to over- or underconfidence \citep{DBLP:conf/icml/GuoPSW17}. In this section, we argue that over- and underconfidence are not just a source of miscalibration, but also a confounder that obscures the intrinsic calibration properties of models and makes it harder to compare across model families.

\textbf{Quantifying confidence.}\quad
Predictions of overconfident models tend to be overly ``peaky'' (low entropy), such that an increase in temperature (positive temperature factor) would be necessary to make them optimally confident, and vice versa for underconfident models. We can therefore quantify confidence in terms of the temperature scaling factor by which the logits of the unscaled model would have to be multiplied to provide optimal confidence. 

The optimal confidence depends on the model \emph{and} the dataset. Ideally, a model would be optimally confident across all distribution shifts, indicating that its confidence is well calibrated to the difficulty of the data. In practice, most models are slightly overconfident in-distribution, and tend to become more overconfident as data moves further from the training distribution (\Cref{fig:app_imagenet_c_with_confidence}, bottom row).

Models can show the opposite trend if they are \emph{under}confident in-distribution. As an example, we include the EfficientNet-NoisyStudent family in \Cref{fig:app_imagenet_c_with_confidence}. These models tend to be underconfident (optimal temperature factor $< 1$; \Cref{fig:app_imagenet_c_with_confidence}, bottom right). Underconfident models may paradoxically show \emph{improved} calibration under distribution shift (lower ECE for higher corruption severities), because their underconfidence balances out the general tendency towards overconfidence on OOD data. However, such underconfident models are not better calibrated in general---they are simply biased towards a high level of distribution shift, and are calibrated worse at weak or no distribution shift. A well-calibrated model should have optimal confidence both in- and out-of-distribution.

\textbf{Normalizing confidence.}\quad
The example of EfficientNet-NoisyStudent illustrates how confidence bias can confound trends in model calibration. This counfounder can be removed by \emph{temperature scaling} \citep{DBLP:conf/icml/GuoPSW17}, i.e. by rescaling model logits to optimize the likelihood on a held-out part of the in-distribution dataset (\ImNetClean in our case). By removing differences in confidence bias between models, temperature scaling reveals a consistent trend for higher calibration error under distribution shift for all models, including EfficientNet-NoisyStudent (\Cref{fig:app_imagenet_c_with_confidence}, second row). Temperature scaling also reveals consistent differences between model families and trends within families for in-distribution calibration (\Cref{fig:imagenet_clean_temp_scaled,fig:imagenet_bit_pretrain}). We therefore study calibration after temperature scaling, in addition to unscaled calibration error and other calibration metrics (\Cref{app:alternative_calibration_metrics}), throughout this work. The benefit of temperature scaling for understanding model calibration is separate from its well-established benefit in reducing calibration error \citep{DBLP:conf/icml/GuoPSW17}. 

\textbf{Label smoothing.}\quad
One method to directly influence the confidence of a model during training is \emph{label smoothing}~\citep{szegedy2016rethinking}. 
In label smoothing, uniformly distributed probability mass is added to the training targets. This decreases the implied confidence of the targets and thus of the model trained on these targets, which can reduce overfitting and improve accuracy.

Label smoothing has been reported to improve calibration \citep{DBLP:conf/nips/MullerKH19}. Here, we argue that label smoothing creates artificially underconfident models and may therefore improve calibration for a specific amount of distribution shift, but does \emph{not} generally improve the intrinsic calibration properties of a model (i.e. its overall calibration across distribution shifts and datasets). 

\begin{figure}[h]
    \centering
    \vspace{20mm}
    \includegraphics[width=0.65\linewidth]{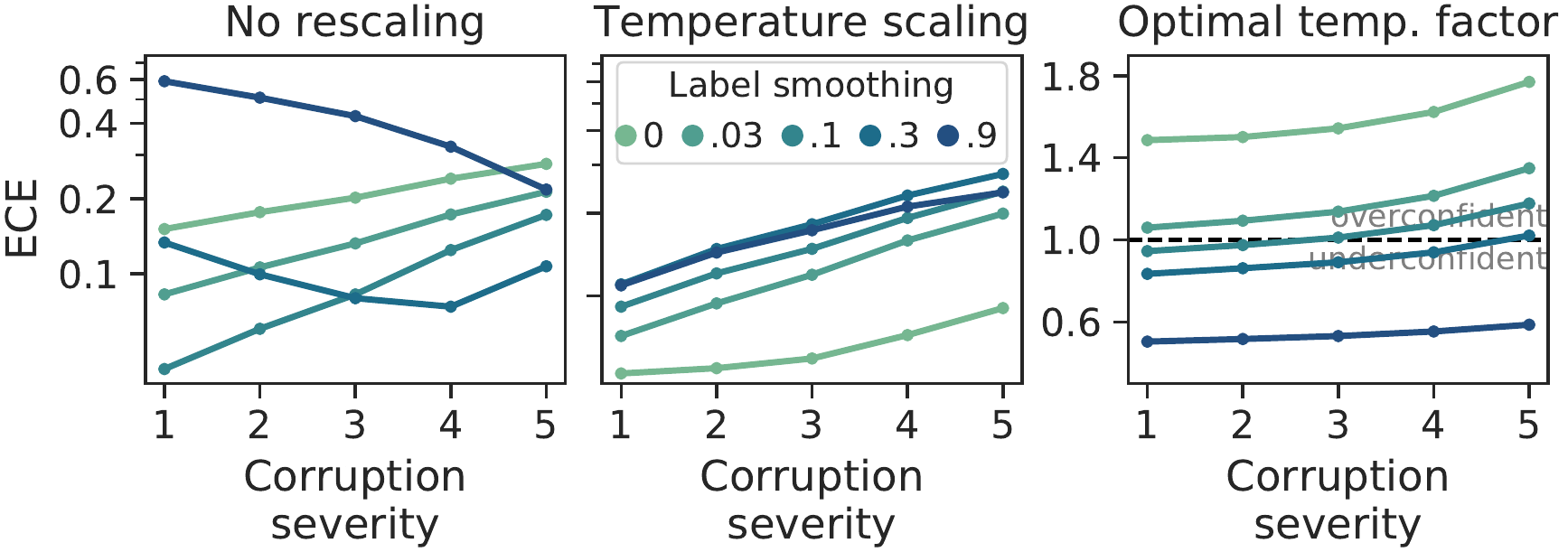}
    \caption{Effect of label smoothing on calibration. EfficientNet-B4 models were trained with the indicated label smoothing on \ImNetClean and evaluated on \ImNetC. Before rescaling, different amounts of non-zero label smoothing appear to yield the best calibration, depending on distribution shift (left). After temperature scaling, it becomes clear that training without label smoothing is best (center). Label smoothing reduces confidence (right). ResNet architectures show similar behavior.}
    \label{fig:app_conf_label_smoothing}
\end{figure}

\Cref{fig:app_conf_label_smoothing} shows the ECE before and after temperature scaling of models trained with different amounts of label smoothing on \ImNetClean and evaluated on \ImNetC. Before temperature scaling (\Cref{fig:app_conf_label_smoothing}, left), the best-calibrated models (lowest ECE) are those trained with label smoothing. Depending on the amount of distribution shift (corruption severity), a different amount of label smoothing is necessary to optimize calibration. After temperature scaling on a held-out part of the \ImNetClean validation set (\Cref{fig:app_conf_label_smoothing}, center), it becomes clear that training without label smoothing actually results in the lowest ECE across all \ImNetC severities. The optimal temperature factor (\Cref{fig:app_conf_label_smoothing}, right) reveals that label smoothing simply biases the model confidence, like temperature scaling, but without targeted optimization. These data suggest that, from a calibration perspective, models should be trained without label smoothing and then recalibrated by temperature scaling \emph{post hoc}.

Label smoothing may explain the anomaly observed for EfficientNet-NoisyStudent under distribution shift (\Cref{fig:app_imagenet_c_with_confidence}, far right). In contrast to all other model families we consider, EfficientNet-NS shows strong \emph{under}confidence before temperature scaling (\Cref{app:model_confidence}); it is also the only model family trained with label smoothing.

\FloatBarrier
\clearpage
\section{ECE Variants} \label{app:ece_variants}
\secsummary{Provides the main plots for a range of alternative ECE estimator variants (e.g. different binning scheme and number of bins).}

\begin{figure*}[p!]
    \vspace{-7mm}
    \newcommand{\figheight}{25mm}
    \newcommand{\figspace}{\vspace{2mm}}
    \centerline{\includegraphics[height=\figheight]{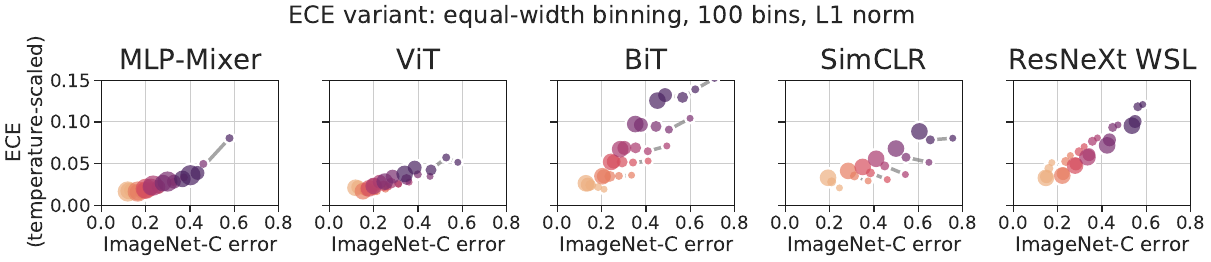}}
    \figspace
    \centerline{\includegraphics[height=\figheight]{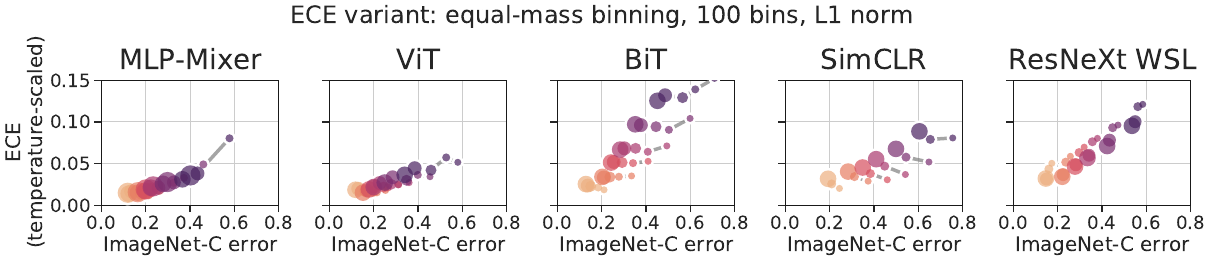}}
    \figspace
    \centerline{\includegraphics[height=\figheight]{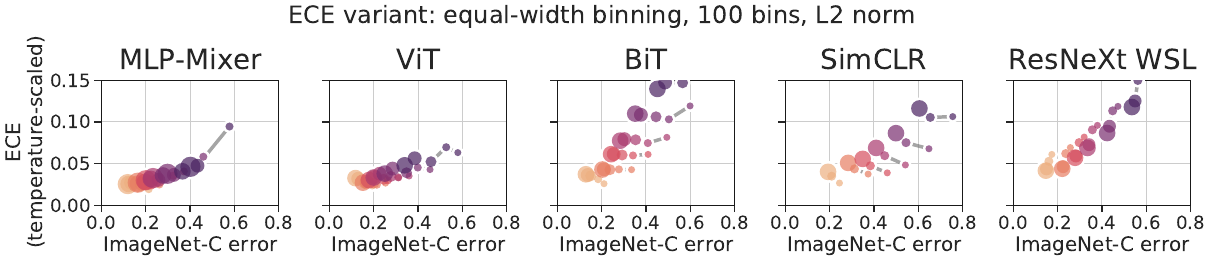}}
    \figspace
    \centerline{\includegraphics[height=\figheight]{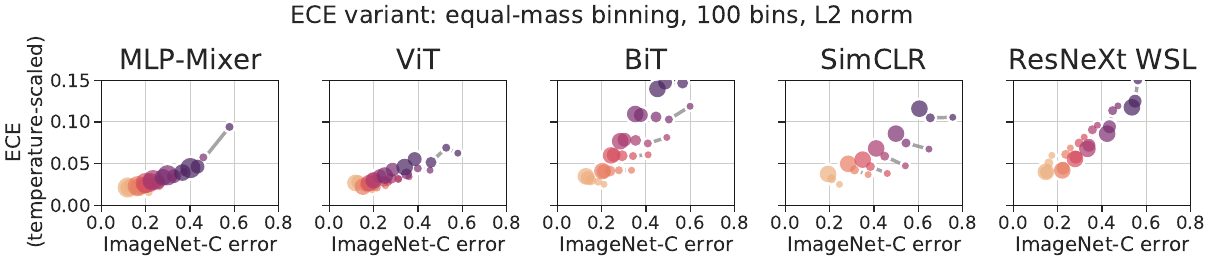}}
    \figspace
    \centerline{\includegraphics[height=\figheight]{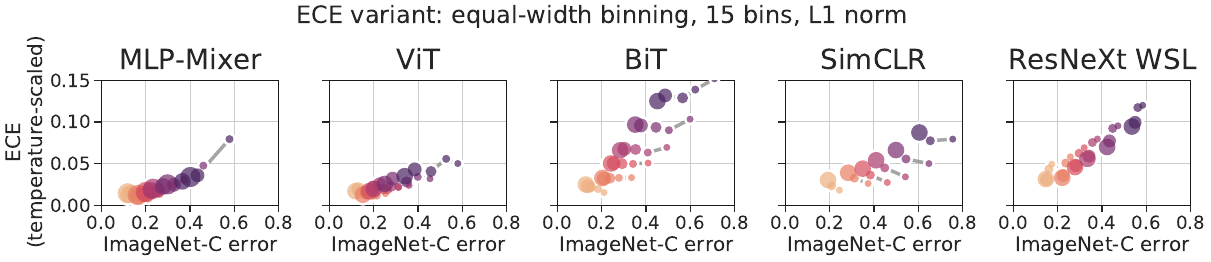}}
    \figspace
    \centerline{\includegraphics[height=\figheight]{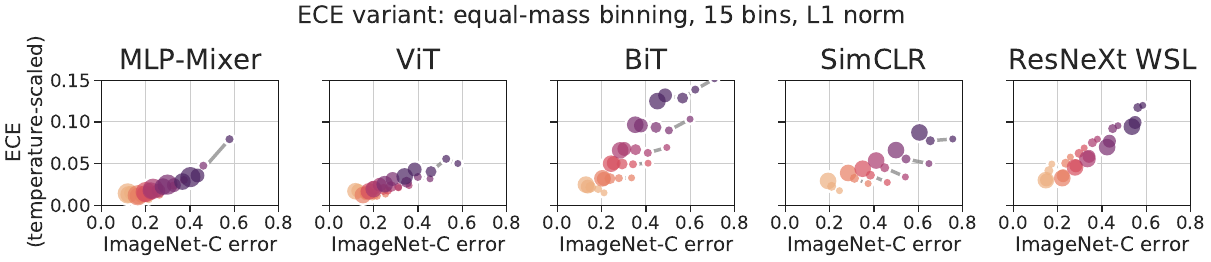}}
    \figspace
    \centerline{\includegraphics[height=\figheight]{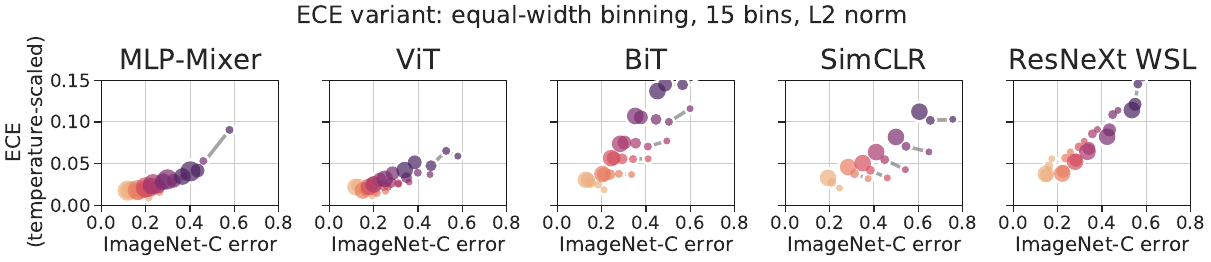}}
    \figspace
    \centerline{\includegraphics[height=\figheight]{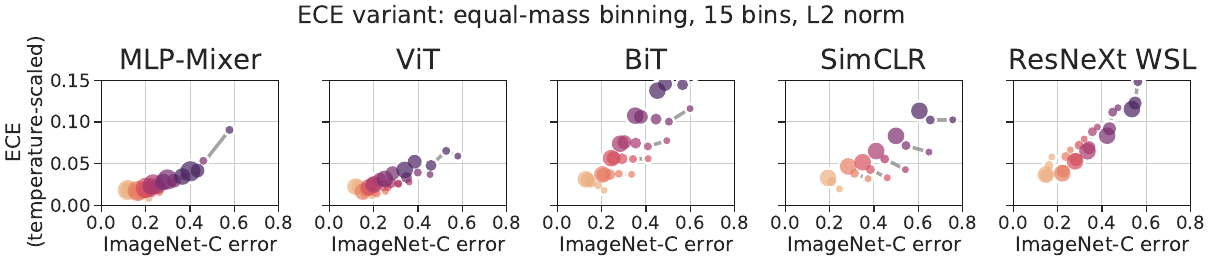}}
    \caption{Related to \Cref{fig:imagenet_c}. Each row shows the calibration and accuracy on \ImNetC as in \Cref{fig:imagenet_c}, bottom row (i.e. after temperature scaling), but for different ECE variants. The variant is indicated in the title of each row. While absolute values can differ between variants, relative relationships between models are robust to the metric variant.}
    \label{fig:imagenet_c_ece_variants}
\end{figure*}

\begin{figure*}[p!]
    \newcommand{\figwidth}{95mm}
    \newcommand{\figspace}{\vspace{10mm}}
    \newcommand{\fighspace}{\hspace{4mm}}
    \centerline{\includegraphics[width=\figwidth,valign=t]{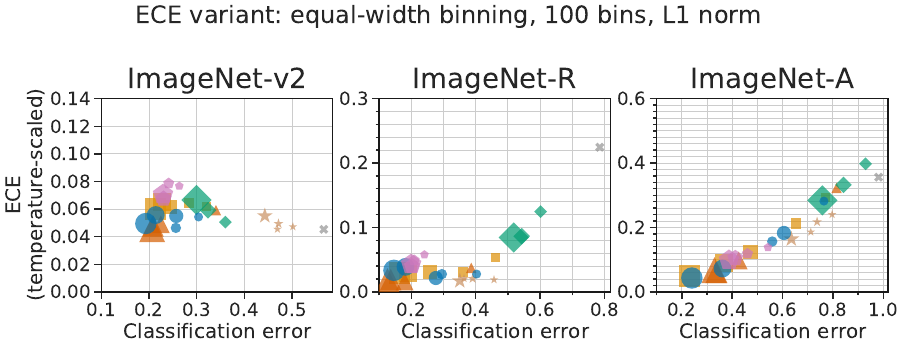}
                \fighspace
                \includegraphics[width=\figwidth,valign=t]{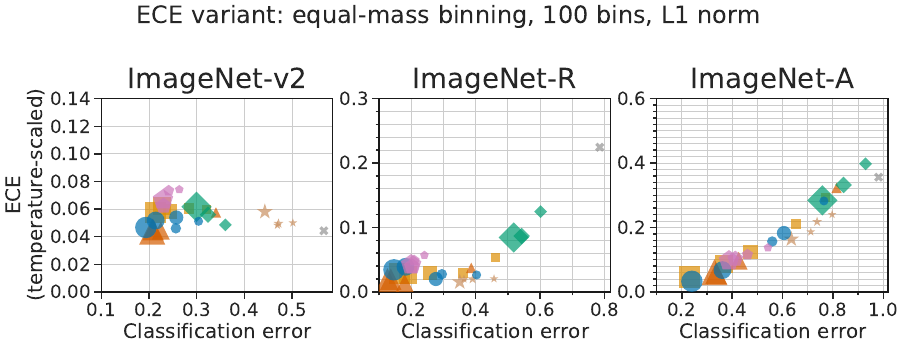}}
    \figspace
    \centerline{\includegraphics[width=\figwidth,valign=t]{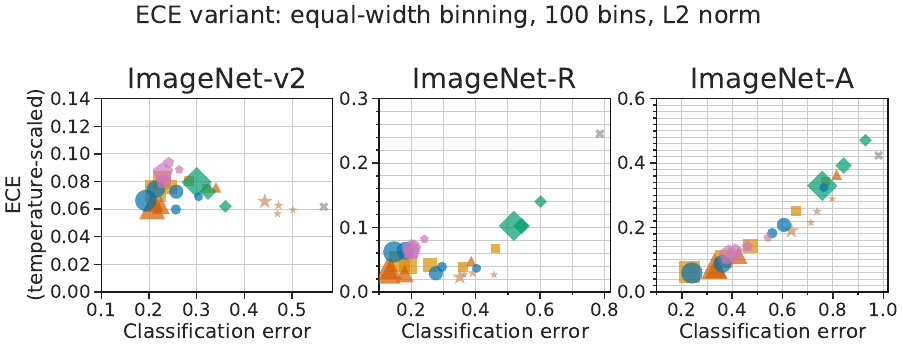}
                \fighspace
                \includegraphics[width=\figwidth,valign=t]{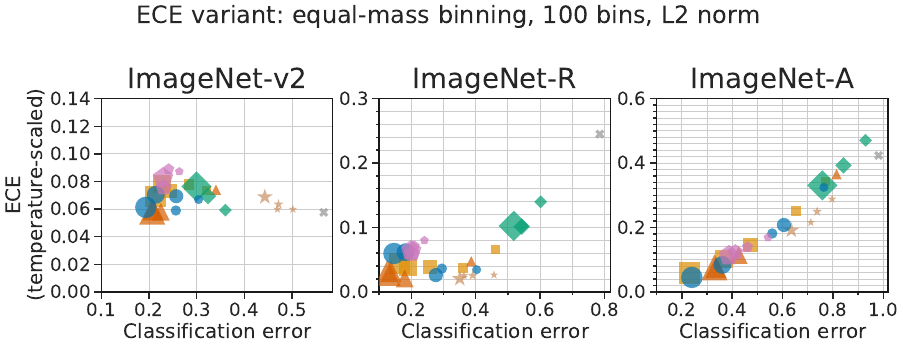}}
    \figspace
    \centerline{\includegraphics[width=\figwidth,valign=t]{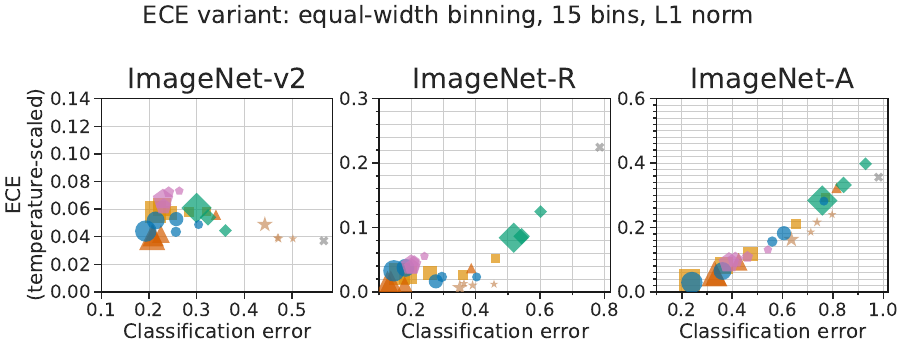}
                \fighspace
                \includegraphics[width=\figwidth,valign=t]{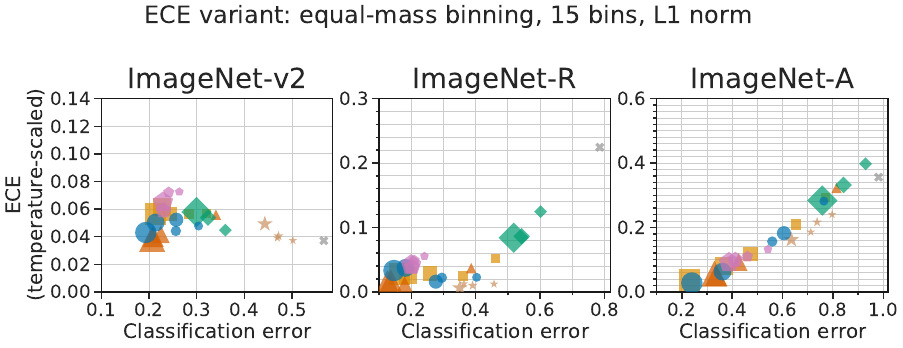}}
    \figspace
    \centerline{\includegraphics[width=\figwidth,valign=t]{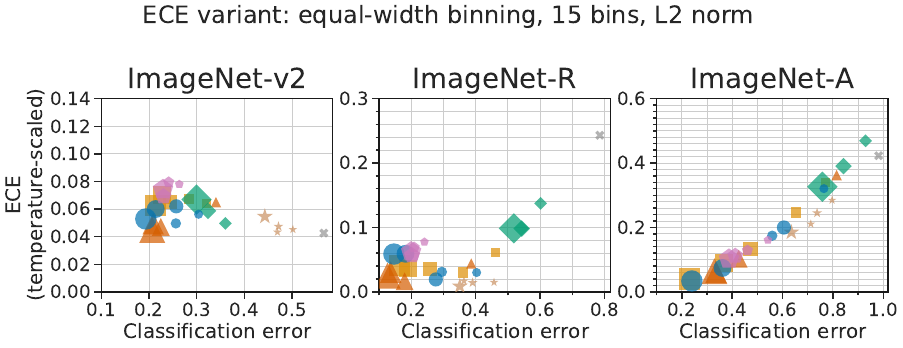}
                \fighspace
                \includegraphics[width=\figwidth,valign=t]{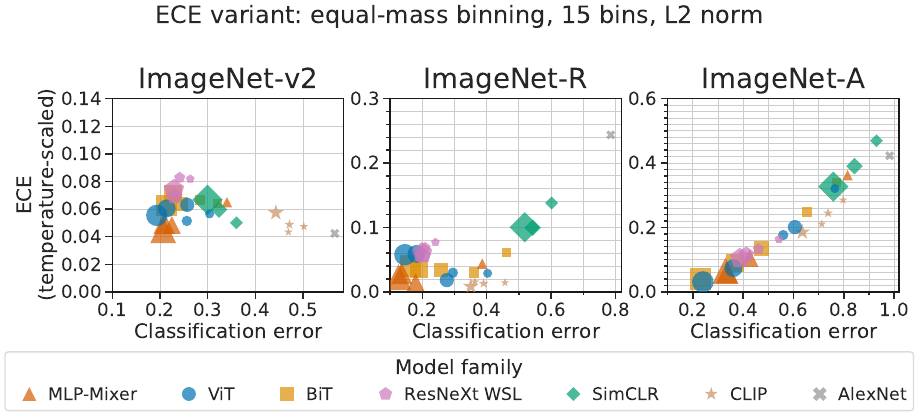}}
    \caption{Related to \Cref{fig:ood_calibration}. Calibration and accuracy on OOD datasets as in \Cref{fig:ood_calibration}, bottom row (i.e. after temperature scaling), but for different ECE variants. The variant is indicated in the title of each set of plots. While absolute values can differ between variants, relative relationships between models are robust to the metric variant.}
    \label{fig:ood_ece_variants}
\end{figure*}

\begin{figure*}[b!]
    \centerline{\includegraphics[width=\textwidth]{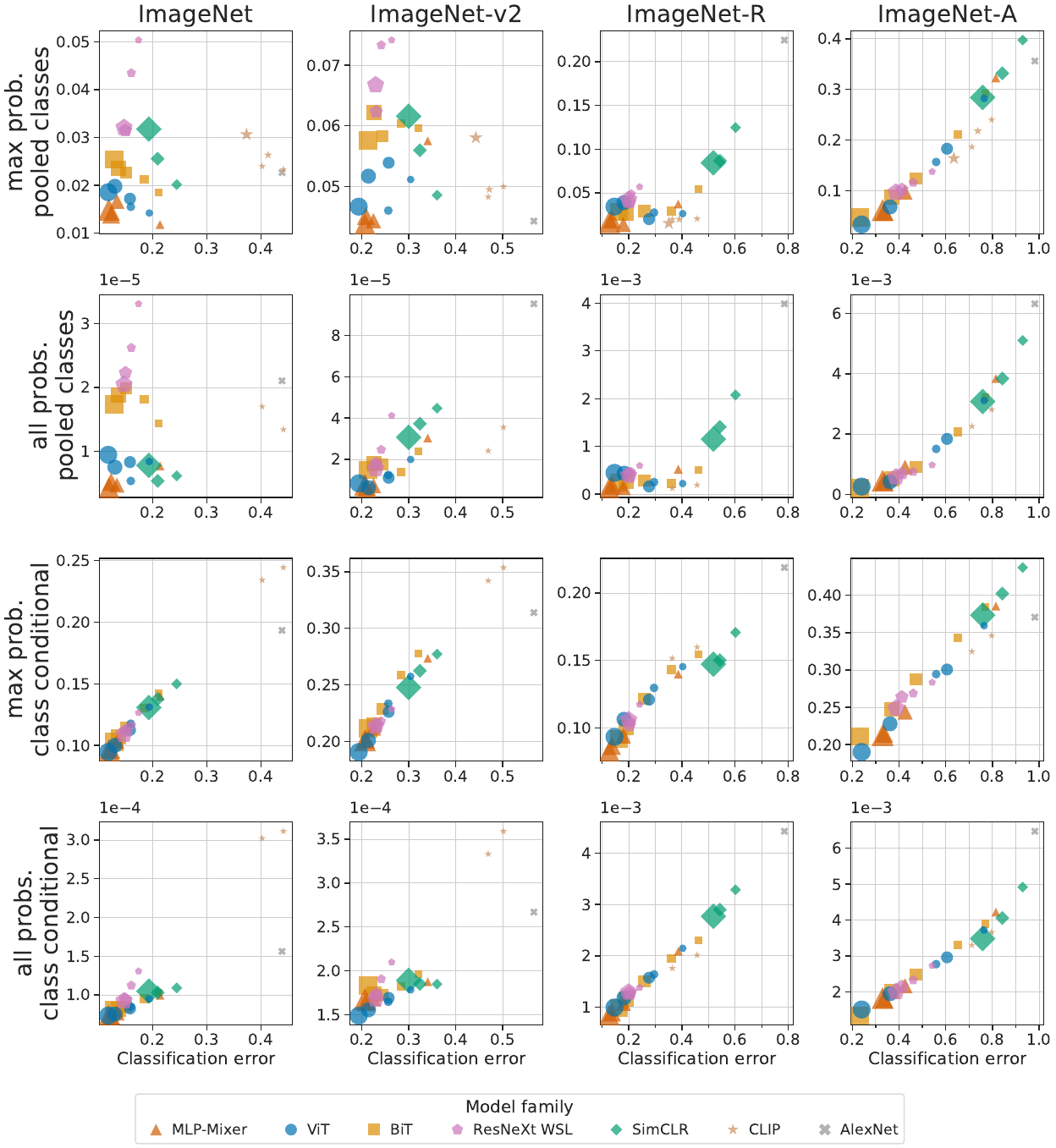}}
    \caption{Further ECE variants (after temperature scaling). The top row shows the variant used in the main paper, which considers only the maximum predicted probability (``top-label calibration'') and pools across classes. The remaining rows show other variants as discussed in \citet{DBLP:conf/cvpr/NixonDZJT19}. L1-normalization and adaptive binning was used in all cases (100 bins for pooled-class metrics; 15 bins for class-conditional metrics). Although the specific rankings between models depend on the ECE variant \citep{DBLP:conf/cvpr/NixonDZJT19}, our main conclusions hold for all variants. Specifically, the same model families tend to be Pareto-optimal across all ECE variants. Also, the relationship between ECE and accuracy is largely consistent across ECE variants.}
    \label{fig:app_alternative_eces_temperature_scaling}
\end{figure*}

As discussed in \Cref{sec:related_work,sec:pitfalls}, while ECE is a well-defined quantity, estimating it requires binning and thus a choice of binning scheme and bin size. In addition, variants of ECE such as root mean squared calibration error (RMSCE, \citealt{DBLP:conf/cvpr/NixonDZJT19}) exist. In RMSCE, the difference between accuracy and confidence in each bin is $\ell_2$-normalized, in contrast to the $\ell_1$-normalization of standard ECE. This causes larger errors to be upweighted in RMSCE. Further ECE variants consider all classes, instead of just the class with the highest predicted probability (top-label), or consider classes independently and report an average of class-wise ECEs. Different ECE variants may rank models differently~\citep{DBLP:conf/cvpr/NixonDZJT19}, which could lead to the conclusion that ECE estimators are fundamentally inconsistent. However, we find that such inconsistencies in model rank are resolved by considering ECE and classification error jointly (\Cref{fig:imagenet_c_ece_variants,fig:ood_ece_variants,fig:app_alternative_eces_temperature_scaling}). While ranks between models may change across ECE variants, these models differ in classification error, such that it is always clear which model is Pareto-optimal in terms of ECE and classification error. For example, for \ImNetVtwo in \Cref{fig:ood_ece_variants}, the ranking of BiT models (orange squares) changes slightly between some of the ECE variants. However, the models differ so much in classification error that the differences in ECE between metric variants are likely irrelevant (also see \Cref{app:abstention}, which shows that differences in classification error typically have a larger influence on decision cost than differences in ECE).

\section{Alternative Calibration Metrics} \label{app:alternative_calibration_metrics}
\secsummary{Provides negative log-likelihood, Brier score, and reliability diagrams for all models and datasets.}

\begin{figure*}[b!]
    \newcommand{\figspace}{\vspace{3mm}}
    \centerline{\includegraphics[width=4.3587in,trim={0mm 8mm 0mm 0mm},clip,right]{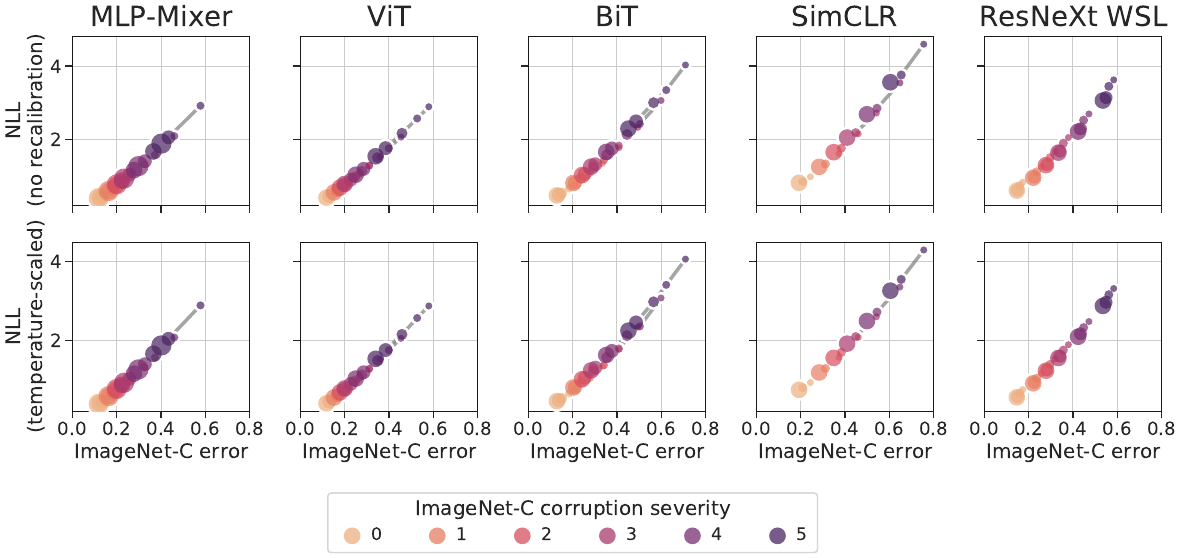}\hspace{25mm}}
    \figspace
    \centerline{\includegraphics[width=4.4712in,trim={0mm 8mm 0mm 3mm},clip,right]{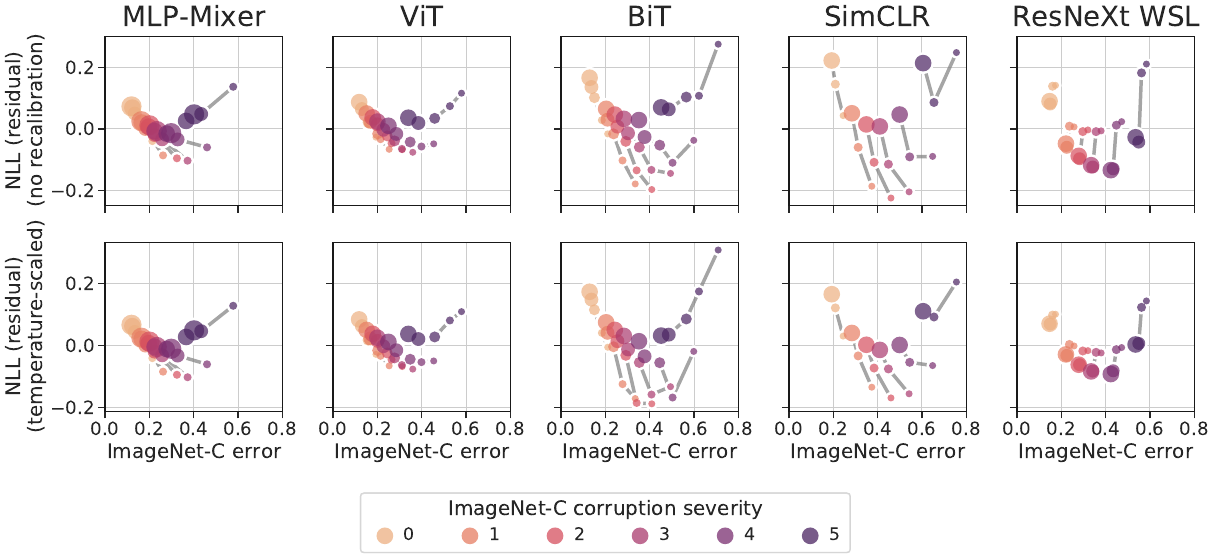}\hspace{25mm}}
    \figspace
    \centerline{\includegraphics[width=4.4541in,trim={0mm 8mm 0mm 3mm},clip,right]{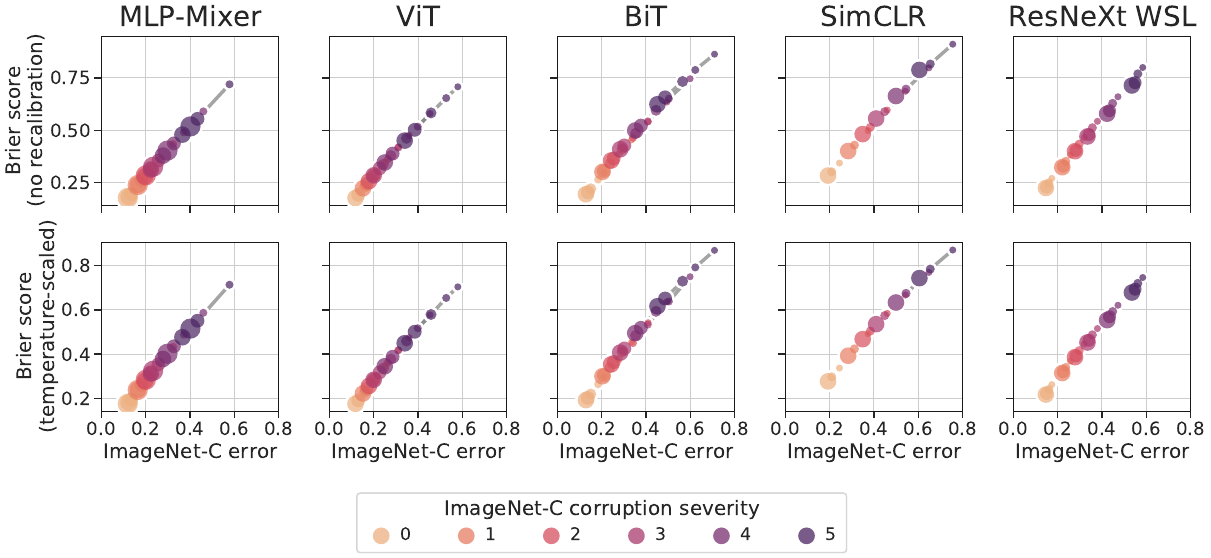}\hspace{25mm}}
    \figspace
    \centerline{\includegraphics[width=4.5099in,trim={0mm 0mm 0mm 3mm},clip,right]{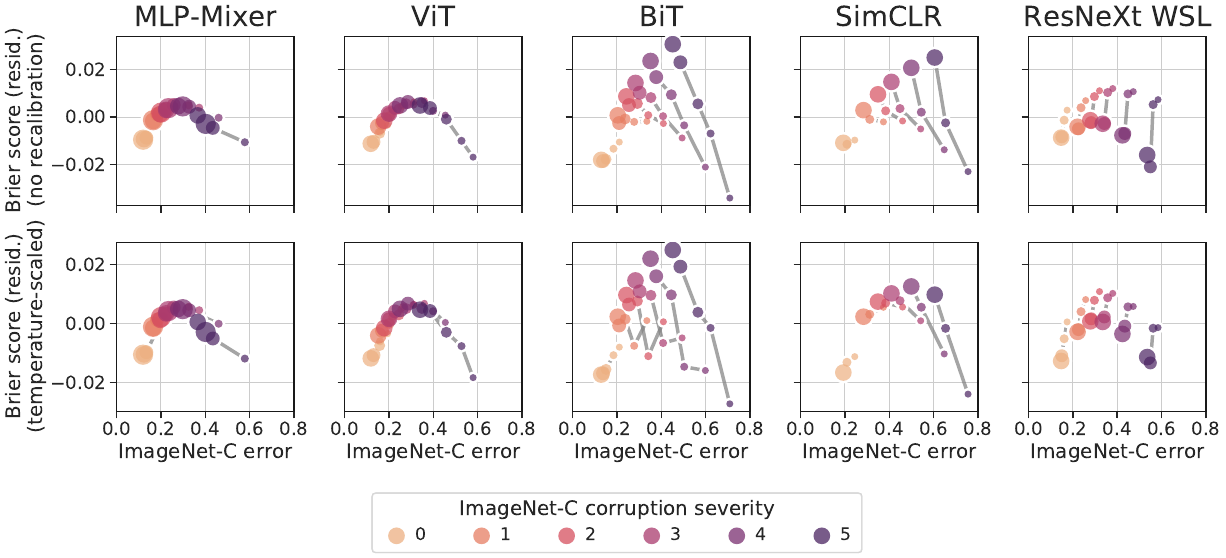}\hspace{25mm}}
    \caption{Alternative calibration metrics for \ImNetC: negative log-likelihood (NLL) and Brier score. Plotted as in \Cref{fig:imagenet_c}. Second and fourth rows show residuals as described in \Cref{fig:alternative_calibration_metrics}.}
    \label{fig:app_alternative_calibration_metrics_imagenet_c}
\end{figure*}

To confirm that our findings are not dependent on our choice of Expected Calibration Error as our main calibration metric, we provide results for two alternative calibration metrics: negative log-likelihood (NLL) and Brier score~\citep{brier1950verification}. \Cref{fig:alternative_calibration_metrics} in the main text covers \ImNetClean, \ImNetVtwo, \ImNetR, and \ImNetA. Results for \ImNetC are provided in \Cref{fig:app_alternative_calibration_metrics_imagenet_c}

\begin{figure*}[p!]
    \newcommand{\figheight}{37mm}
    \newcommand{\mixerfigheight}{40mm}  
    \newcommand{\figspace}{\vspace{5mm}}
    \centerline{\includegraphics[height=\mixerfigheight]{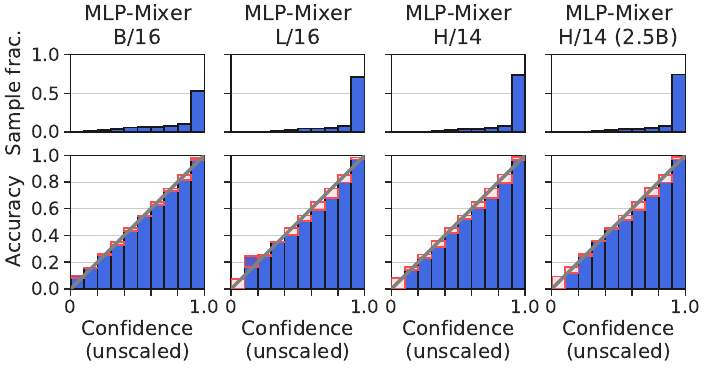}}
    \figspace
    \centerline{\includegraphics[height=\figheight]{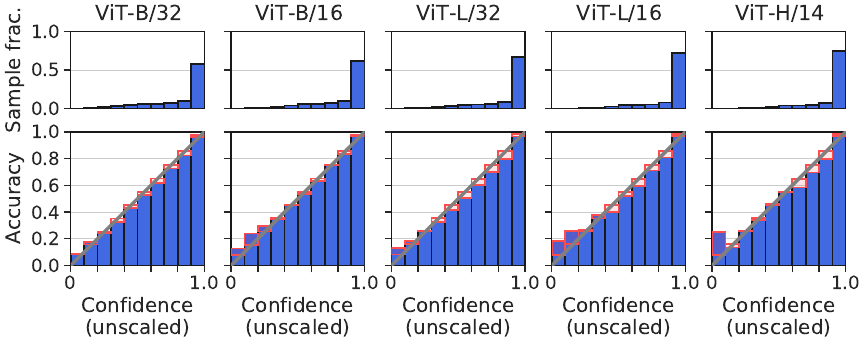}}
    \figspace
    \centerline{\includegraphics[height=\figheight]{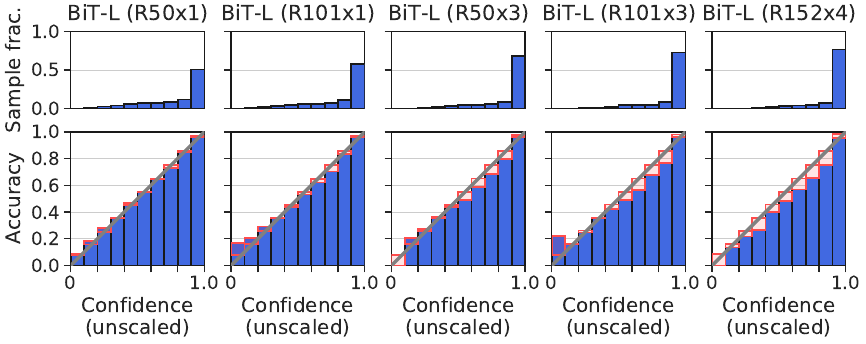}}
    \figspace
    \centerline{\includegraphics[height=\figheight]{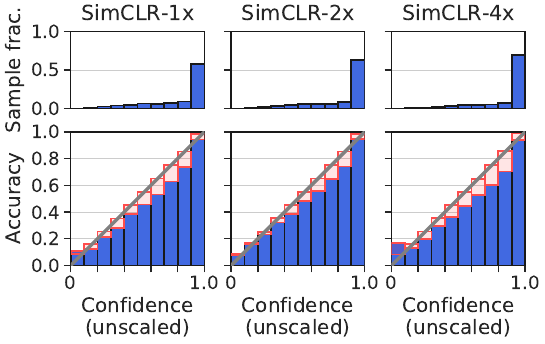}
                \hspace{10mm}
                \includegraphics[height=\figheight]{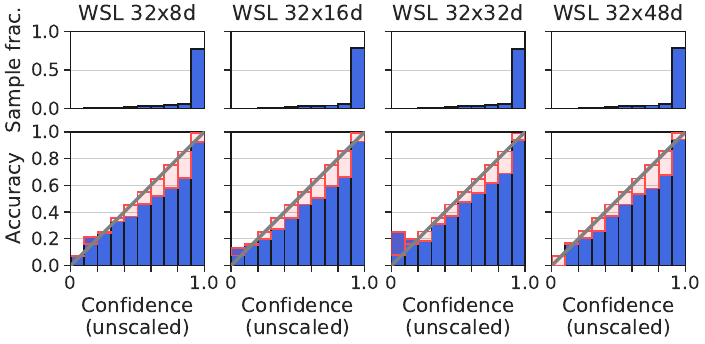}}
    \figspace
    \centerline{\includegraphics[height=\figheight]{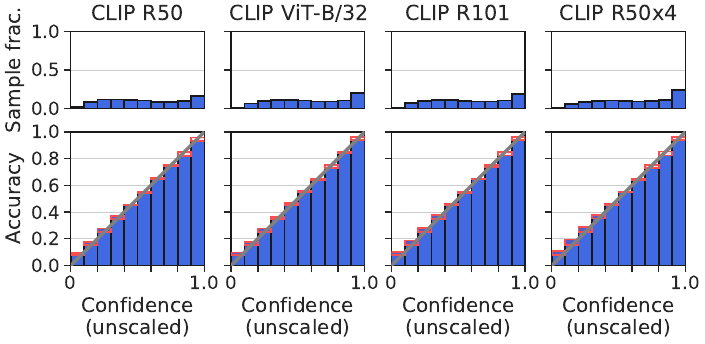}
                \hspace{10mm}
                \includegraphics[height=\figheight]{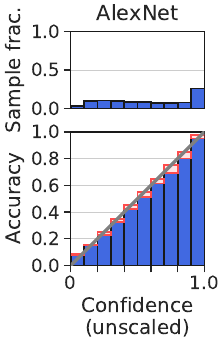}}
    \caption{Reliability diagrams on \ImNetClean for all models, before temperature scaling. Red boxes indicate the error compared to perfect calibration. The histogram at the top shows the distribution of confidence values for the dataset.}
    \label{fig:reliability_diagrams_unscaled}
\end{figure*}

\begin{figure*}[p!]
    \newcommand{\figheight}{37mm}
    \newcommand{\mixerfigheight}{40mm}  
    \newcommand{\figspace}{\vspace{5mm}}
    \centerline{\includegraphics[height=\mixerfigheight]{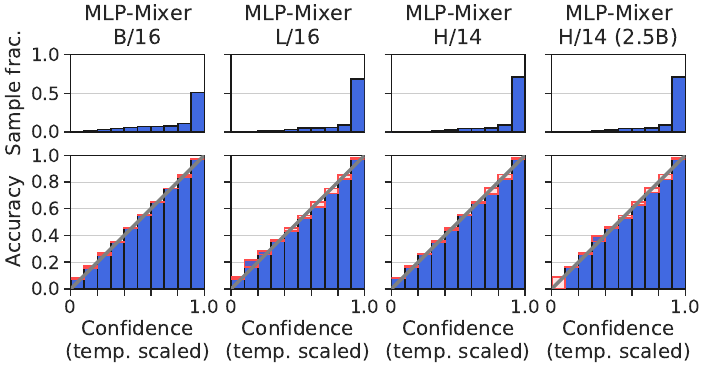}}
    \figspace
    \centerline{\includegraphics[height=\figheight]{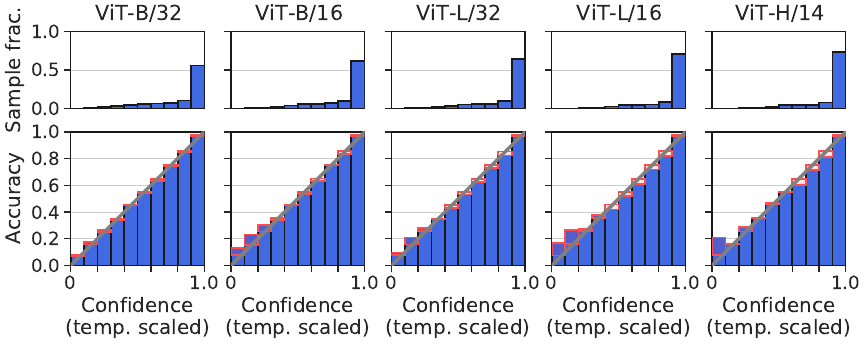}}
    \figspace
    \centerline{\includegraphics[height=\figheight]{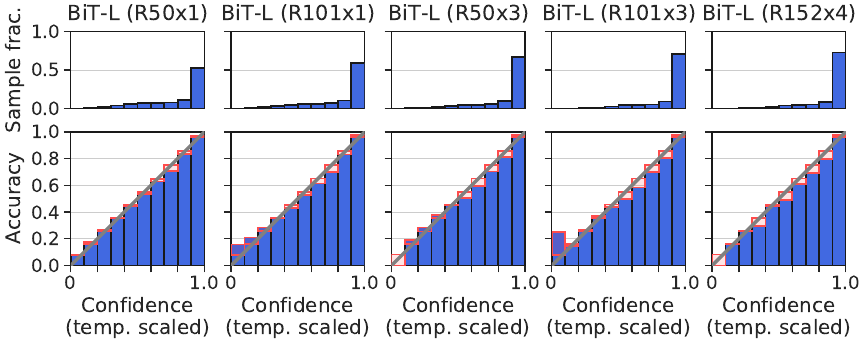}}
    \figspace
    \centerline{\includegraphics[height=\figheight]{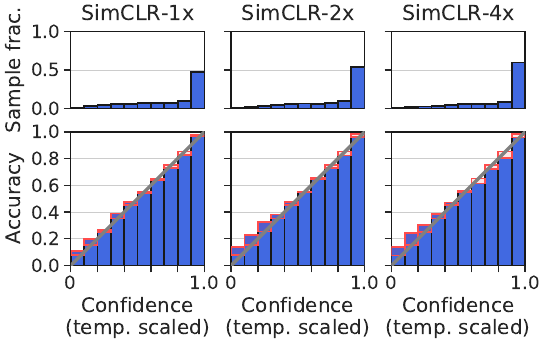}
                \hspace{10mm}
                \includegraphics[height=\figheight]{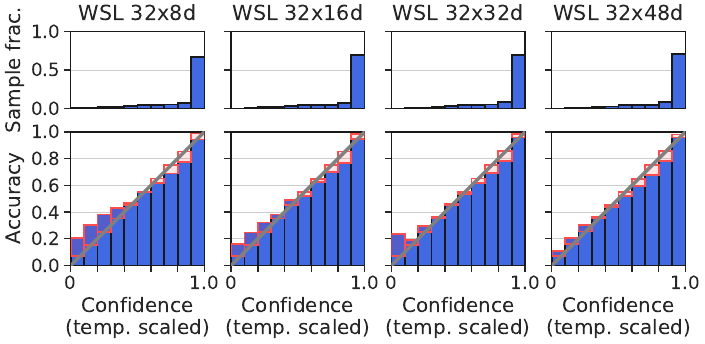}}
    \figspace
    \centerline{\includegraphics[height=\figheight]{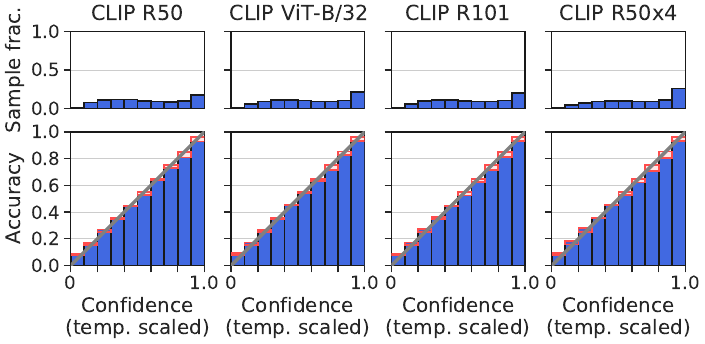}
                \hspace{10mm}
                \includegraphics[height=\figheight]{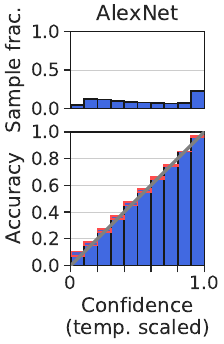}}
    \caption{Reliability diagrams on \ImNetClean for all models, after temperature scaling. Red boxes indicate the error compared to perfect calibration. The histogram at the top shows the distribution of confidence values for the dataset.}
    \label{fig:reliability_diagrams_temp_scaled}
\end{figure*}

Furthermore, we provide reliability diagrams~\citep{degroot1983comparison} on \ImNetClean for all models, both before (\Cref{fig:reliability_diagrams_unscaled}) and after (\Cref{fig:reliability_diagrams_temp_scaled}) temperature scaling. These diagrams visualize model calibration across the whole confidence range, rather than summarizing calibration into a scalar value.

\vspace{5mm}
\textit{Figures for \Cref{app:ece_variants} and \Cref{app:alternative_calibration_metrics} are on the following pages.}

\clearpage
\thispagestyle{empty}

\clearpage
\thispagestyle{empty}

\end{document}